\documentclass[twocolumn]{bmcart}

\usepackage{amsthm,amsmath, multirow}

\RequirePackage{hyperref}
\usepackage[utf8]{inputenc} 

\usepackage{graphicx}
\usepackage[strings]{underscore}
\graphicspath{ {./images/} }

\startlocaldefs
\endlocaldefs
\usepackage{xcolor}
\newcommand{\rev}[1]{\textcolor{black}{#1}}

\usepackage{textpos}

\begin{document}

\begin{frontmatter}

\begin{fmbox}
\hfill\textsf{\small Accepted for publication by Journal of Cardiovascular Magnetic Resonance}
\dochead{Research}


\title{Automated Quality Control in Image Segmentation: Application to the UK Biobank Cardiac MR Imaging Study}


\author[
   addressref={aff1},                   	
   corref={aff1},                       	
   noteref={n1},                        	
   email={r.robinson16@imperial.ac.uk}   	
]{\inits{RR}\fnm{Robert} \snm{Robinson}}
\author[
   addressref={aff1},
   email={v.valindria15@imperial.ac.uk}
]{\inits{VV}\fnm{Vanya V} \snm{Valindria}}
\author[
   addressref={aff1},
   email={w.bai@imperial.ac.uk}
]{\inits{WB}\fnm{Wenjia} \snm{Bai}}
\author[
   addressref={aff1},
   email={o.oktay13@imperial.ac.uk}
]{\inits{OO}\fnm{Ozan} \snm{Oktay}}
\author[
   addressref={aff1},
   email={b.kainz@imperial.ac.uk}
]{\inits{BK}\fnm{Bernhard} \snm{Kainz}}
\author[
   addressref={aff2},
   email={h.suzuki@imperial.ac.uk}
]{\inits{HS}\fnm{Hideaki} \snm{Suzuki}}
\author[
   addressref={aff4, aff5},
   email={m.sanghvi@qmul.ac.uk}   
]{\inits{MMS}\fnm{Mihir M.} \snm{Sanghvi}}
\author[
   addressref={aff4, aff5},
   email={n.aung@qmul.ac.uk}
]{\inits{NA}\fnm{Nay} \snm{Aung}}
\author[
   addressref={aff4},
   email={j.paiva@qmul.ac.uk}
]{\inits{JMP}\fnm{José Miguel} \snm{Paiva}}
\author[
   addressref={aff4, aff5},
   email={f.zemrak@qmul.ac.uk}
]{\inits{FZ}\fnm{Filip} \snm{Zemrak}}
\author[
   addressref={aff4, aff5},
   email={kenneth.fung@bartshealth.nhs.uk}
]{\inits{KF}\fnm{Kenneth} \snm{Fung}}
\author[
   addressref={aff6},
   email={elena.lukaschuk@cardiov.ox.ac.uk}
]{\inits{EL}\fnm{Elena} \snm{Lukaschuk}}
\author[
   addressref={aff4, aff5},
   email={a.lee@qmul.ac.uk}
]{\inits{AML}\fnm{Aaron M.} \snm{Lee}}
\author[
   addressref={aff6},
   email={valentina@simula.no}
]{\inits{VC}\fnm{Valentina} \snm{Carapella}}
\author[
   addressref={aff6, aff7},
   email={dryj@yuhs.ac}
]{\inits{YJK}\fnm{Young Jin} \snm{Kim}}
\author[
   addressref={aff6},
   email={stefan.piechnik@cardiov.ox.ac.uk}
]{\inits{SKP}\fnm{Stefan K.} \snm{Piechnik}}
\author[
   addressref={aff6},
   email={stefan.neubauer@cardiov.ox.ac.uk}
]{\inits{SN}\fnm{Stefan} \snm{Neubauer}}
\author[
   addressref={aff4, aff5},
   email={s.e.petersen@qmul.ac.uk}
]{\inits{SEP}\fnm{Steffen E.} \snm{Petersen}}
\author[
   addressref={aff3},
   email={chris.s.page@gsk.com}
]{\inits{CP}\fnm{Chris} \snm{Page}}
\author[
   addressref={aff2, aff8},
   email={p.matthews@imperial.ac.uk}
]{\inits{PM}\fnm{Paul M} \snm{Matthews}}
\author[
   addressref={aff1},
   email={d.rueckert@imperial.ac.uk}
]{\inits{DR}\fnm{Daniel} \snm{Rueckert}}
\author[
   addressref={aff1},
   email={b.glocker@imperial.ac.uk}
]{\inits{BG}\fnm{Ben} \snm{Glocker}}


\address[id=aff1]{
  \orgname{Biomedical Image Analysis Group, Department of Computing, Imperial College London}, 
  \street{Queen's Gate},                     %
  \postcode{SW7 2AZ},                                
  \city{London},                              
  \cny{UK}                                    
}
\address[id=aff2]{%
  \orgname{Division of Brain Sciences, Dept. of Medicine, Imperial College London},
  \street{Queen's Gate},                     %
  \postcode{SW7 2AZ},                          
  \city{London},                              
  \cny{UK}                                    
}
\address[id=aff3]{%
  \orgname{GlaxoSmithKline Research and Development},
  \street{Stockley Park},                     %
  \postcode{UB11 1BT},                          
  \city{Uxbridge},                              
  \cny{UK}                                    
}
\address[id=aff4]{%
  \orgname{William Harvey Research Institute, NIHR Barts Biomedical Research Centre, Queen Mary University of London},
  \street{Charterhouse Square},                     %
  \postcode{EC1M 6BQ},                          
  \city{London},                              
  \cny{UK}                                    
}
\address[id=aff5]{%
  \orgname{Barts Heart Centre, Barts Health NHS Trust},
  \street{West Smithfield},                     %
  \postcode{EC1A 7BE},                          
  \city{London},                              
  \cny{UK}                                    
}
\address[id=aff6]{%
  \orgname{Division of Cardiovascular Medicine, Radcliffe Department of Medicine, University of Oxford},
  \postcode{OX3 9DU},                          
  \city{Oxford},                              
  \cny{UK}                                    
}
\address[id=aff7]{%
  \orgname{Department of Radiology, Severance Hospital, Yonsei University College of Medicine},
  \city{Seoul},                              
  \cny{South Korea }                                    
}
\address[id=aff8]{%
  \orgname{UK Dementia Research Institute, Imperial College London},
  \street{Queen's Drive},                     %
  \postcode{SW7 2AZ},
  \city{London},                              
  \cny{UK}                                    
}

\begin{artnotes}
\note[id=n1]{Email addresses of all authors:
RR: r.robinson16@imperial.ac.uk,
VV: v.valindria15@imperial.ac.uk,
WB: w.bai@imperial.ac.uk, 
OO: o.oktay13@imperial.ac.uk,
BK: b.kainz@imperial.ac.uk,
HS: h.suzuki@imperial.ac.uk,
MMS: m.sanghvi@qmul.ac.uk,
NA: n.aung@qmul.ac.uk,
JMP: j.paiva@qmul.ac.uk,
FZ: filip@zemrak.co.uk,
KF: k.fung@qmul.ac.uk,
EL: elena.lukaschuk@cardiov.ox.ac.uk,
AML: a.lee@qmul.ac.uk,
VC: valentina@simula.no,
YJK: dryj@yuhs.ac,
SKP: stefan.piechnik@cardiov.ox.ac.uk,
SN: stefan.neubauer@cardiov.ox.ac.uk,
SEP: s.e.petersen@qmul.ac.uk,
CP: chris.s.page@gsk.com,
PMM: p.matthews@imperial.ac.uk,
DR: d.rueckert@imperial.ac.uk,
BG: b.glocker@imperial.ac.uk.
}

\end{artnotes}



\begin{abstractbox}

\begin{abstract}
\parttitle{Background}
The trend towards large-scale studies including population imaging poses new challenges in terms of quality control (QC). This is a particular issue when automatic processing tools such as image segmentation methods are employed to derive quantitative measures or biomarkers for further analyses. Manual inspection and visual QC of each segmentation result is not feasible at large scale. However, it is important to be able to automatically detect when a segmentation method fails in order to avoid inclusion of wrong measurements into subsequent analyses which could otherwise lead to incorrect conclusions.
\parttitle{Methods}
To overcome this challenge, we explore an approach for predicting segmentation quality based on Reverse Classification Accuracy, which enables us to discriminate between successful and failed segmentations on a per-cases basis. We validate this approach on a new, large-scale manually-annotated set of 4,800 cardiac magnetic resonance scans. We then apply our method to a large cohort of 7,250 cardiac MRI on which we have performed manual QC. 
\parttitle{Results}
We report results used for predicting segmentation quality metrics including Dice Similarity Coefficient (DSC) and surface-distance measures. As initial validation, we present data for 400 scans demonstrating 99\% accuracy for classifying low and high quality segmentations using the predicted DSC scores. As further validation we show high correlation between real and predicted scores and 95\% classification accuracy on 4,800 scans for which manual segmentations were available. We mimic real-world application of the method on 7,250 cardiac MRI where we show good agreement between predicted quality metrics and manual visual QC scores.
\parttitle{Conclusions}
We show that RCA has the potential for accurate and fully automatic segmentation QC on a per-case basis in the context of large-scale population imaging as in the UK Biobank Imaging Study.

\end{abstract}




\end{abstractbox}
\end{fmbox}

\end{frontmatter}




\section*{Background}
Biomedical image data are increasingly processed with automated image analysis pipelines which employ a variety of tools to extract clinically useful information. It is important to understand the limitations of such pipelines and assess the quality of the results being reported. This is a particular issue when we consider large-scale population imaging databases comprising thousands of images such as the UK Biobank (UKBB) Imaging Study \cite{Sudlow2015}. There are often many modules in automated pipelines \cite{Shariff2010} where each may contribute to inaccuracies in the final output and reduce the overall quality of the analysis, e.g. intensity normalisation, segmentation, registration and feature extraction. On a large scale, it is infeasible to perform a manual, visual inspection of all outputs, and even more difficult to perform quality control (QC) within the pipeline itself. We break down this challenge and focus on the automated QC of image segmentation.

Image segmentation is the process of partitioning an image into several parts where each of these parts is a collection of pixels (or voxels) corresponding to a particular structure. The purpose of segmentation is to derive quantitative measures of these structures, e.g. calculating ventricular volume or vessel thickness. Automated segmentation is desired to reduce workload for this tedious, time-consuming and error prone task. A number of these methods have been developed, ranging from basic region-growing techniques and graph cuts to more advanced algorithms involving machine learning \cite{DeBruijne2016} and, more recently, Deep Learning in the form of Convolutional Neural Networks (CNNs) \cite{Bai2017}.

Segmentation performance is traditionally evaluated on a labelled validation dataset, which is a subset of the dataset that is the algorithm does not see during training. This evaluation is done using a series of metrics to compare the predicted segmentation and a reference `ground truth' (GT). Popular metrics include volumetric overlap \cite{Crum2006}, surface distances or other statistical measures \cite{Taha2015}. Due to the lack of actual GT, manual expert annotations are used as reference, despite inter- and intra-rater variability. Once a segmentation method is deployed in clinical practice no such quantitative evaluation can be carried out routinely.

Evaluating the average performance of an algorithm on validation data is arguably less important than being able to assess the quality on a per-case basis, and it is crucial to identify cases where the segmentation has failed. We show that we can effectively predict the per-case quality of automated segmentations of 3D cardiac MRI (CMR) from the UKBB which enables fully automated QC in large-scale population studies and clinical practice.

In this article we will first present related work that attempts to address the problem of automated QC at large-scale. Our method and datasets are then described in detail before we present our results and discuss their implications.

\subsection*{Related work}
Despite its practical importance, there is relatively little work on automatically predicting performance of image analysis methods. Much of the prior work on automated quality control has focused on the quality of images themselves. This focus on image quality assessment (IQA) is also true in the medical-imaging community \cite{Carapella2016a,Zhang_2016}. In the context of image segmentation, there exist only a few methods outlined here.

Algorithms often rely on `labels' to support their training. In our case, each label would indicate the quality of each segmentation, either by categorical label, e.g. 0 for `poor' and 1 for `good', or by continuous value such as a Dice Similarity Coefficient. In cases where such labelled data is scarce, Reverse Validation \cite{Zhong2010} and Reverse Testing \cite{Fan2006} use labels generated by one model, trained on a subset of available data, to train another model which is evaluated on the remaining data. This is effectively cross-validation where the amount of labelled data is limited. In Reverse Testing, `some rules' are created to assess the performance of and rank the different models. In our context, this would involve creating a segmentation quality model from a subset of MR scans, and their corresponding segmentations, which can then be tested on the remaining images. Different models would be created and tested in order to choose the best model. The difficulty in these methods is that we require all of the scans to be accurately segmented in order to train, and to evaluate, a good model. That is, we need a large, fully-annotated training dataset which is often not available in our field. Additionally, Reverse Validation and Reverse Testing do not allow us to identify individual cases where a segmentation may have failed; instead they focus upon the segmentation method as a whole.

In a method proposed by Kohlberger et al., the quality of segmentations is assessed on a per-case basis using machine learning. The group used 42 different hand-crafted statistics about the intensity and appearance of multi-organ computed-tomography (CT) scans to inform their model. Whilst this method achieved good performance metrics and an accuracy of around 85\%, it requires a lot of training data with both good and bad segmentations which is non-trivial to obtain.

In this work, we adopt the recently-proposed approach of Reverse Classification Accuracy (RCA) \cite{Valindriaa}. Unlike Reverse Validation and Reverse Testing, RCA can accurately predict the quality of a segmentation on a case-by-case basis only requiring a relatively small set of accurately segmented reference images. In RCA, the predicted segmentation being assessed is used to create a small model to re-segment the reference images for which segmentations are available. If at least one image in the reference set is re-segmented well, the predicted segmentation, that we wish to assess, must have been of good quality. We employ RCA to perform segmentation quality analysis on a per-case basis while only requiring a small set of reference images and segmentations.


\section*{Methods and Data}
\label{sec:method}

Our purpose is to have a system that is able to predict the per-case quality of a segmentation produced by any algorithm deployed in clinical practice. We want our method to not only give us a prediction of the quality of the segmentation, but to be able to identify if that segmentation has failed. To this end, we employ RCA which will give a prediction about the quality of individual segmentations.

\subsection*{Reverse Classification Accuracy}

In RCA the idea is to build a model, also known as an `RCA classifier', solely using one test image and its predicted segmentation which acts as pseudo ground truth. This classifier is then evaluated on a reference dataset for which segmentations are available. There are two possible outcomes to this procedure:
\begin{itemize}
\item \textbf{Case 1:} assuming that the predicted segmentation is of \textit{good quality}, the created model should be able to segment at least one of the reference images with high accuracy. This is likely to be a reference image which is similar to the test image.
\item \textbf{Case 2:} if none of the reference images are segmented successfully, then the predicted segmentation is likely to be of \textit{poor quality}.
\end{itemize}
These assumptions are valid if the reference dataset is representative of the test data. This is usually the case in the context of machine learning where the reference data could have been used in the first place to train the automated method for which we want to predict test performance. If the test data were very different, the automated method would in any case not perform well, and RCA scores would reflect this. It is a great advantage that the same reference dataset can be used to train an automated segmentation method, and also afterwards serves as the reference database enabling prediction of performance after deployment of the segmentation method.

The performance of the RCA classifier on the reference set is measured with any chosen quality metric, e.g., the Dice similarity coefficient (DSC). The highest score among all reference images determines the quality \textit{estimate} for the predicted segmentation obtained for a test image.

The original work on RCA \cite{Valindriaa} explored a variety of possible classifiers that could be trained on a single test-image and its segmentation including Atlas Forests (AF) \cite{Zikic2014} and Convolutional Neural Networks (CNNs). In this context, and throughout this paper, an `atlas' refers to an image-segmentation pair whose segmentation has been verified by a manual annotator. In Valindria's paper, a simple single-atlas registration classifier outperformed both the AF and CNN approaches in predicting segmentation accuracy. For this reason, we chose to use this simple approach for the model in our work. Registration is the process of aligning two or more images based upon similar content within them, e.g. structures or intensities. Rigid registration restricts the images to move only by linear translations and rotations. More complex non-rigid registration methods exists that allow for differences in scale between the images and for more complex distortions. The single-atlas registration classifier in RCA works by performing non-rigid registration of the test-image to a set of individual reference-images. The resulting transformations are then used to warp the test-segmentation. This yields a set of warped segmentations which are quantitatively compared to the reference segmentations. The overlap between the pairs is calculated as the Dice Similarity Coefficient (DSC) whilst boundary agreement is computed using surface-distance metrics. The best metric values among the reference set are taken to be the prediction for the quality of the test-segmentation.

\begin{figure}[tbh!]
    \includegraphics[width=0.45\textwidth]{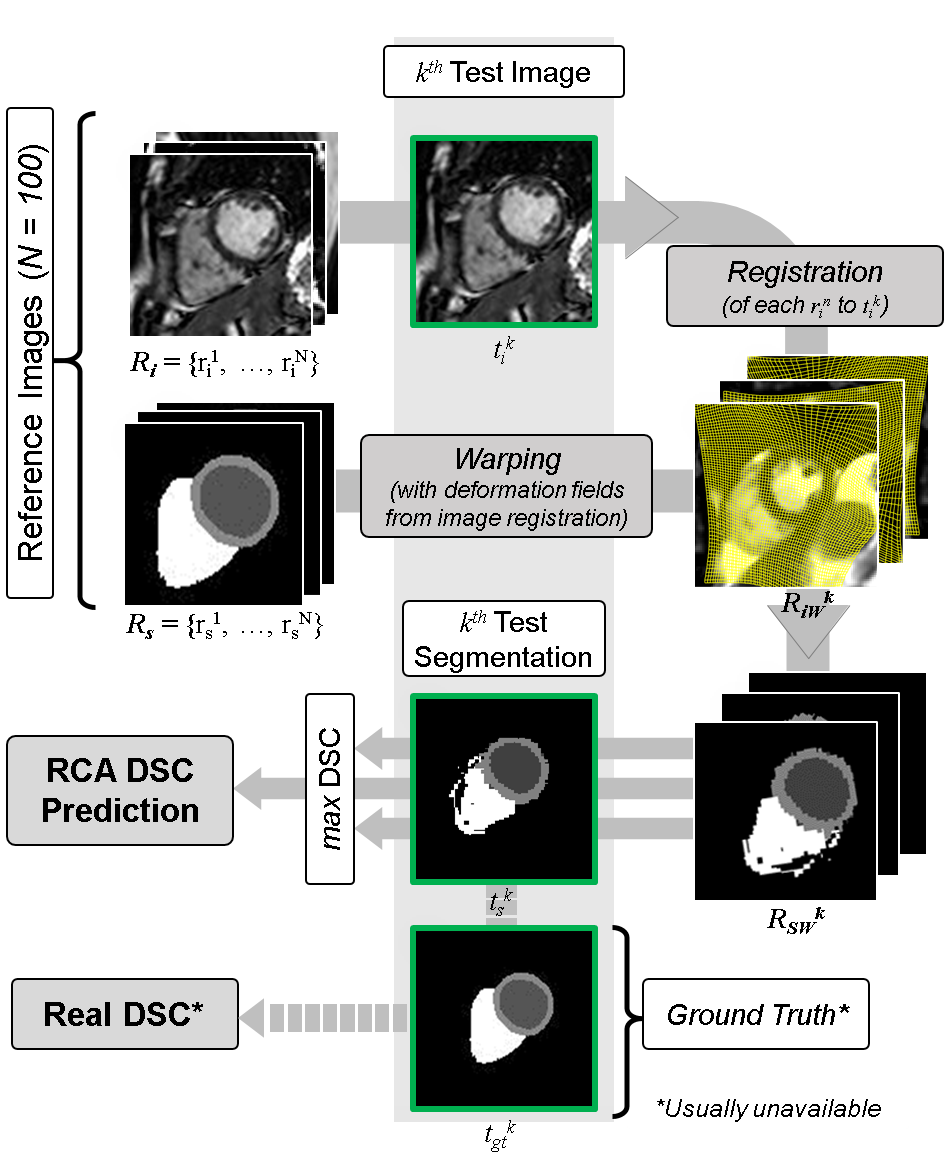}
	\caption{\csentence{Reverse Classification Accuracy - Single-atlas Registration Classifier.}
		Reverse Classification Accuracy (RCA), with single-atlas registration classifier, as applied in our study. A set of reference images are first registered to the test-image before the resulting transformations are used to warp the corresponding reference segmentations. Dice Similarity Coefficient (DSC) is calculated between the warped segmentations and the test-segmentation with the maximum DSC taken as a proxy for the accuracy of the test-segmentation. Note that in practice, the ground truth test-segmentation is absent. Images and segmentation annotated as referred to in the text}
		\label{fig:RCAsingleatlas}
\end{figure}

We chose to modify the single-atlas registration classifier from that used in Valindria et al.'s proposal of the RCA method \cite{Valindriaa}. Processing and modifying the test-segmentation is not usually desirable as this may introduce discretization artefacts adding false-positives into the binary labelmap. We choose to perform the single-atlas registration in reverse: we register the reference-images to the test-image and use this transformation to warp the reference segmentations. This results in a set of warped segmentations in the test-image space which are then compared to the test-segmentation. Figure~\ref{fig:RCAsingleatlas} gives an overview of RCA as applied in our study. We now set out our framework more formally.

For the RCA reference images, we use a set $ \mathbf{R}_{\mathrm{i}} = \left \{ r_{\mathrm{i}}^{1}, \cdots, r_{\mathrm{i}}^{N} \right \}$ of $N$ cardiac atlases with reference segmentations $ \mathbf{R}_{\mathrm{s}} = \left \{ r_{\mathrm{s}}^{1}, \cdots, r_{\mathrm{s}}^{N} \right\}$. We have a test set $\mathbf{T}_{\mathrm{i}} = \left \{ t_{\mathrm{i}}^{1}, \cdots, t_{\mathrm{i}}^{M} \right\}$ of $M$ images with automatically generated predicted segmentations $\mathbf{T}_{s} = \left \{ t_{\mathrm{s}}^{1}, \cdots, t_{\mathrm{s}}^{M} \right\}$ whose quality we would like to assess. If the GT segmentations $ \mathbf{T}_{\mathrm{gt}} = \left \{ t_{\mathrm{gt}}^{1}, \cdots, t_{\mathrm{gt}}^{M} \right\}$ for $\mathbf{T}_{\mathrm{i}}$ exist, one can evaluate the accuracy of these quality assessments. Using RCA we estimate the quality of those predicted segmentations and compare the estimates to the real quality with respect to the GT. 

In the case where $m = k$, we take the $k$\textsuperscript{th} test image $t_{\mathrm{i}}^{k}$ and its predicted segmentation $t_{\mathrm{s}}^{k}$. To apply RCA, all reference images $\mathbf{R}_{\mathrm{i}}$ are first registered to $t_{\mathrm{i}}^{k}$ by performing a rigid registration in the form of a centre of mass (CoM) alignment. Initial versions of our work \cite{Robinson2017} used landmark registration \cite{Oktay2017} at this stage, but we now opt for CoM alignment to reduce computational cost. We then perform non-linear registration of each aligned reference image in $\mathbf{R}_{\mathrm{i}}$ to the test image to get warped reference images $\mathbf{R}_{\mathrm{iW}}^{k}$. The same transformations are used to warp the GT reference segmentations $ \mathbf{R}_{\mathrm{s}}$ to get the set $\mathbf{R}_{\mathrm{sW}}^{k}$. For each warped segmentation in $\mathbf{R}_{\mathrm{sW}}^{k}$ we compare against the predicted segmentation $t_{\mathrm{s}}^{k}$ by evaluating a set of metrics detailed below. The best value for each metric over all warped reference segmentations is taken to be the \textit{prediction} of segmentation accuracy for $t_{\mathrm{s}}^{k}$. In our validation studies, we can compute the \textit{real} metrics by comparing the predicted segmentation $t_{\mathrm{s}}^{k}$ with its GT $t_{\mathrm{gt}}^{k}$.

\subsection*{Evaluation of Predicted Accuracy}
The segmentation quality metrics predicited with RCA include the Dice similarity coefficient (DSC), mean surface distance (MSD), root-mean-square surface distance (RMS) and Hausdorff distance (HD). For two segmentations, $\textbf{A}$ and $\textbf{B}$, DSC is a measure of overlap given by $\mathrm{DSC} = 2 \left|\textbf{A} \cap \textbf{B} \right| / \left( |\textbf{A}| + |\textbf{B}|\right)$. The surface distance between a point $a$ on the surface of \textbf{A} and the surface of \textbf{B} is given by the minimum of the euclidean norm $\min_{b \in B} \left|\left| a - b \right|\right|_{2}$ for all points $b$ in the surface of \textbf{B}. The total surface distance is the sum of the surface distances for all points in \textbf{A}. We don't assume symmetry in these calculations, so the surface distance is also calculated from \textbf{B} to \textbf{A}. By taking the mean over all points we get the MSD. RMS is calculated by taking the square of surface distances, averaging and taking the square root. Finally, the HD is taken to be the maximum surface distance.

For each test image, we report the evaluation metrics for each class label: left-ventricular (LV) cavity, LV myocardium (LVM) and right-ventricular (RV) cavity (RVC). We incorporate the voxels of the papillary muscles into the LV cavity class. The right-ventricular myocardium is difficult to segment because it is thin, therefore it is seldom seen in SAX CMR segmentations and not considered in this paper. For each evaluation metric (DSC and surface distances), we could report two difference average values: either a whole-heart average by combining all class labels into a single `whole-heart' (WH) class or, second, by taking the mean across the individual class scores. The WH-class average is usually higher because a voxel attributed to an incorrect class will reduce the mean calculated across the classes, but will actually be considered correct in the single WH-class case.

\subsection*{Experimental Setup}
We perform three investigations in this work which are summarised in Table~\ref{tab:summary}: A) an initial small-scale validation study on 400 test contours of 80 images from an internal cardiac atlas dataset; B) a large-scale validation study on another 4,805 UKBB image with manual ground truth and C) a real-world application to a large set of 7,250 UKBB 3D CMR segmentations.

\subsubsection*{Reference Dataset, $N=100$}
The reference image set is the same in all of our studies. We use 100 2D-stack short-axis (SA) end-diastolic (ED) CMR scans that were automatically segmented and validated by expert clinicians at Hammersmith Hospital, London. Note that the reference set is distinct from all other datasets used. Compared with data from the UKBB, the reference set are of higher in-plane resolution at $1.25 \times 1.25$ mm and have a smaller slice thickness of 2 mm. These images are not used for any purpose other than for this reference set. When choosing a reference set, one should ensure that it is representative of the dataset on which it is being used i.e. it should be of the same domain (SAX CMR in this case) and large enough to capture some variability across the dataset. A reference set that is too small may underestimate the RCA prediction. Though we argue that this may be better than overestimating the quality of a segmentation. Conversely, too large a reference set will cause a significant lengthening of RCA execution time. \rev{We have explored the effect of the RCA reference set size on the prediction accuracy as part of our evlauation which we present in our Discussion.}

\subsubsection*{Experiment A: Initial Validation Study, $N=400$}
\textbf{Data:} We validate RCA on predicting cardiac image segmentation quality using 100 manually verified image-segmentation pairs (different from the reference dataset). Each atlas contains a SA ED 3D (2D-stack) CMR and its manual segmentation. The images have a pixel-resolution of $1.25 \times 1.25 \times 2.0$ mm and span $256 \times 256 \times 56$ voxels. Each manual segmentation identifies voxels belonging to the LV cavity, LV myocardium and RV cavity separating the heart from the background class.

For validation, we generate automatic segmentations of our atlases with varying quality. We employ Random Forests (RFs) with $T=500$ trees and a maximum depth of $D=40$ trained on the same set of 100 cardiac atlases used for testing RCA in this experiment. RFs allow us to produce a variety of test segmentations with intentionally degraded segmentation quality by limiting the depth of the trees during test time. We obtain 4 sets of 100 segmentations by using depths of 5, 20, 30 and 40. Thus, a total of 400 segmentations are used in our initial validation study.

\textbf{Evaluation:} We perform RCA on all 400 segmentations to yield predictions of segmentation quality. The manual segmentations allow us to evaluate the real metrics for each automated segmentation. We compare these to the quality predicted by RCA. To identify individual cases where segmentation has failed, we implement a simple classification strategy similar to that in Valindria's work \cite{Valindriaa}. We consider a 2-group binary classification where DSC scores in the range [0.0 0.7) are considered `poor' and in the range [0.7 1.0] are `good'. These boundaries are somewhat arbitrary and would be adjusted for a particular use-case. Other strategies could be employed on a task-specific basis, e.g. formulation as outlier detection with further statistical measures. The thresholding approach allows us to calculate true (TPR) and false (FPR) positive rates for our method as well as an overall accuracy from the confusion matrix.

\subsubsection*{Experiment B: Large-scale Validation on Manually-segmented UKBB Data, $N=4,805$}
In this experiment we demonstrate that RCA is robust for employment in large-scale studies, and indeed produces accurate predictions of segmentation quality on an individual basis. As part of our collaboration under UK Biobank Application 2964 we have access to 4,805 CMR images with manually drawn contours.

\textbf{Data:} In the context of UKBB data, 3D means a stack of 2D acquisitions with slice-thickness of 8.0 mm and slice-gap of 2 mm \cite{Petersen2016}. The CMR scans have in-plane resolution of $1.83 \times 1.83$ mm and span around $192 \times 208$ pixels per slice. The number of slices per scan varies between 4-14 with the majority (89\%) having 9-12 slices.

Petersen and colleagues \cite{Petersen2017,Petersen2017b} manually segmented all slices of each 3D cardiac MRI scan available under the data access application. Several annotators were employed following a standard operating procedure to generate almost 5,000 high-quality segmentations. With these manual segmentations acting as GT we directly compare predicted segmentation scores with real scores at large scale.

We use the same RF trained in Experiment A to perform automated segmentations at various tree depths chosen randomly across the 4,805 scans yielding segmentations of varying quality. In addition to our RF segmentations, we evaluate RCA with 900 segmentations generated with a recent deep learning based approach. As part of the UKBB Application 2964, Bai et al. \cite{Bai2017} have trained a CNN on 3,900 manually segmented images. The remaining 900 were then automatically segmented using the trained network. The results of Bai et al.'s CNN approach reflect the state-of-the-art in automated 3D cardiac MR segmentation with an accuracy matching the performance human experts \cite{Bai2017}.

\textbf{Evaluation:} We perform RCA on all 4,805 RF segmentations to yield predictions of segmentation quality. We also perform RCA separately on the 900 CNN segmentations produced by a state-of-the-art deep learning approach. With the availability of GT manual segmentations, we can evaluate this experiment in the same way as Experiment A.

\subsubsection*{Experiment C: Automatic Quality Control in the UKBB Imaging Study, $N=7,250$}
Having evaluated RCA in the previous two experiments, this experiment mimics how our approach would behave in a real-world application where the GT is unavailable. We apply RCA to segmentations of CMR images from the UKBB.

\textbf{Data:} In total, 7,250 cardiac MR images were available to us through the UKBB resource. Each image has been automatically segmented using a multi-atlas segmentation approach \cite{Bai2013}. As part of a genome-wide association study (GWAS), each automatic segmentation has been checked manually to confirm segmentation quality. As there is no GT segmentation, we rely on manual QC scores for these segmentations assessed by a clinical expert. The manual QC is based only on visual inspection of the basal, mid and apical layers. For each layer a score between 0 and 2 is assigned based on the quality of only the LV myocardium segmentation. The total QC score is thus between 0 and 6, where a 6 would be considered as a highly accurate segmentation. Scores for individual layers were not recorded. Where the UKBB images had a poor field-of-view (FOV), the segmentations were immediately discarded for use in the GWAS study: we have given these images a score of -1. For the GWAS study, poor FOV meant any image in which the entire heart was not visible. We expect that despite the poor FOV of these images, the segmentations themselves may still be of good quality as the algorithms can still see most of the heart. Out of the 7,250 segmented images, 152 have a bad FOV ($\text{QC}=-1$) and 42 have an obviously poor segmentation ($\text{QC}=0$). There are 2, 14, 44, 300, 2866 and 3830 images having QC scores 1 to 6 respectively. This investigation explored how well RCA-based quality predictions correlate with those manual QC scores.

\textbf{Evaluation:} We perform RCA on all 7,250 segmentations to yield predictions of segmentation quality for the LVM. With the absence of GT segmentations, we are unable to perform the same evaluation as in Experiments A and B. In this case, we determine the correlation between the predicted scores from RCA (for LV myocardium) and the manual QC scores. A visual inspection of individual cases is also performed at quality categories.

\begin{table}[h!]
  \caption{\textbf{A summary of the experiments performed in this study.} Experiment A uses data from an internal dataset which is segmented with a multi-atlas segmentation approach and manually validated by experts at Hammersmith Hospital, London. These manual validations are counted as `ground truth’ (GT) and 100 of them are taken for the reference set used in all experiments. UKBB datasets are shown with their application numbers. In experiment C we segment with both random forests (RF) and a convolutional neural network (CNN). In C the CNN from Bai \cite{Bai2017} is used.} \label{tab:summary}
  \begin{tabular}{ccccc}
  \hline 
  \textbf{Experiment}  		& \textbf{Dataset} 				& \textbf{Size} 		& \textbf{GT} 		& \textbf{Seg. Method} 	\\	\hline		
  A 		& Hammersmith   & 100		& Yes		& RF 			\\
  B 		& UKBB-2964	    & 4,805		& Yes		& RF and CNN  	\\
  C 		& UKBB-18545	& 7,250		& No		& Multi-Atlas	\\   \hline
  \end{tabular}
\end{table}


\section*{Results}\label{sec:results}
Here we present results from our three investigations: (A) the initial small-scale validation study; (B) application to a large set of UKBB cardiac MRI with visual QC scores; and (C) a further large-scale validation study on UKBB with manual expert segmentations.

\begin{figure*}[tbh!]
    \includegraphics[width=0.95\textwidth]{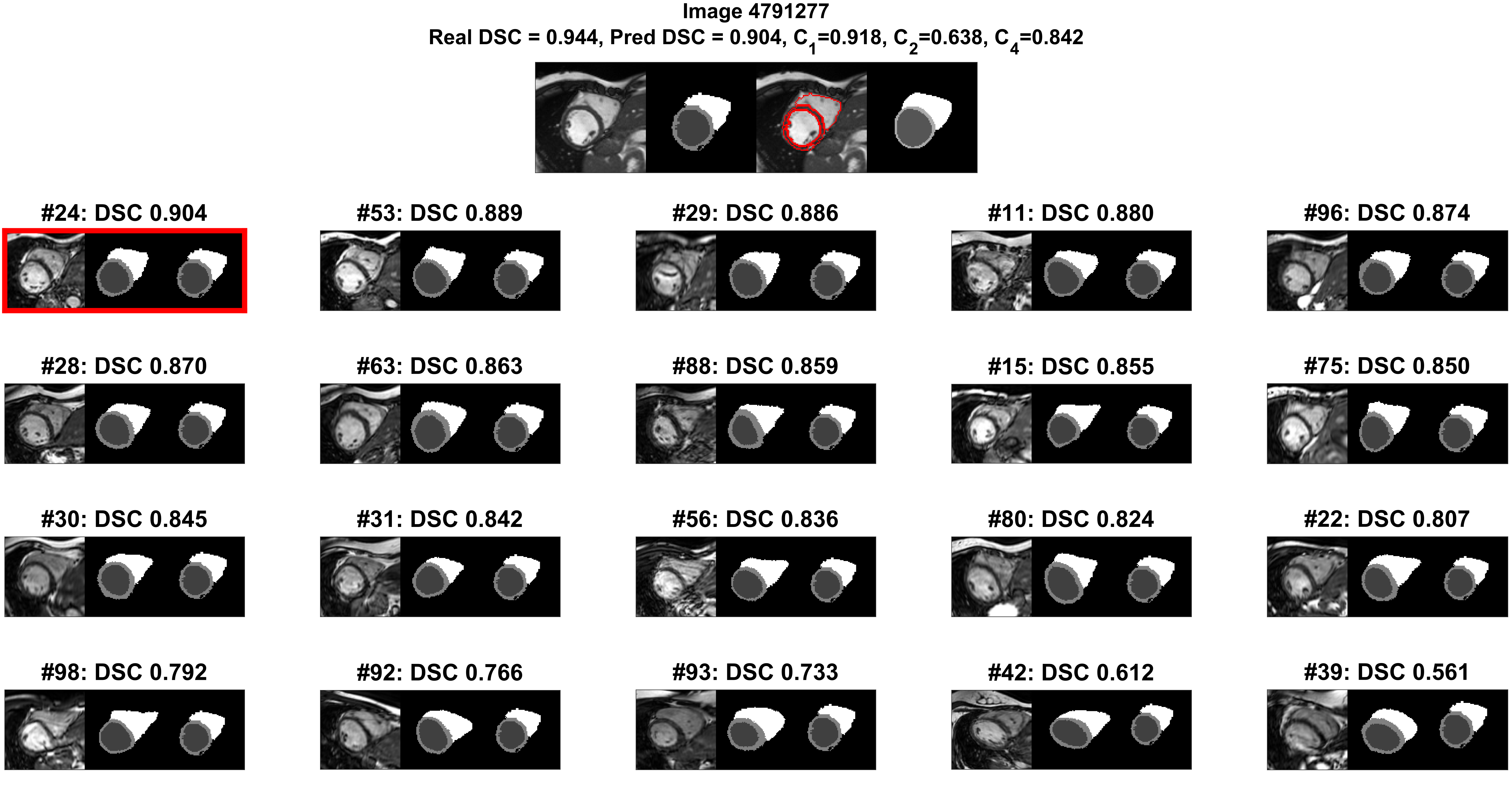}
	\caption{\csentence{Example Results from RCA.} Examples of RCA results on one proposed segmentation. The panels in the top row show (left to right) the MRI scan, the predicted segmentation, an overlay and the manual annotation. The array below shows a subset of the 100 reference images ordered by Dice similarity coefficient (DSC) and equally spaced from highest to lowest DSC. The array shows (left) the reference image, (middle) its ground truth segmentation and (right) the test-segmentation from the upper row which has been warped to the reference image. The real DSC between each reference image and warped segmentation is shown for each pair. RCA-predicted and real GT-calculated DSCs are shown for the whole-heart binary classification case at the top alongside the metrics for each individual class in the segmentation.}
		\label{fig:atlasexamples}
\end{figure*}

Quantitative results for the experiments are presented in each section. Figure~\ref{fig:atlasexamples} demonstrates additional qualitative inspection that is performed on a per-case basis during RCA. The top row of Figure~\ref{fig:atlasexamples} shows the mid-ventricular slice of an ED CMR scan and a RF-generated segmentation which is under test. An overlay of the two are also shown alongside the manual reference segmentation which is not available in practice. Below this, an array of further panels is shown. Each of these panels presents one of the 100 reference images used, its corresponding reference segmentation and the result of warping the segmentation-under-test (top-panel, second image) to this reference image. The calculated DSC between the reference image's GT and the warped segmentation is displayed above each panel. The array shows the reference image with the highest (top-left) and the lowest (bottom-right) calculated DSC with the remaining panels showing DSCs that are uniformly spaced amongst the remaining 98 reference images. We can see in this example that there is a large range of predicted DSC values, but only the maximum prediction, selected in red, is used as the prediction of segmentation quality. For the example in Figure~\ref{fig:atlasexamples} we show a `good' quality segmentation-under-test for which we are predicting a DSC of 0.904 using RCA. The real DSC between the segmentation-under-test and the GT manual segmentation is 0.944. Note that in this case these values are calculated for the `whole-heart' where individual class labels are merged into one. These values are shown above the top panel along with the DSC calculated on a per-class basis.

For considerations of space, we do not show more visual examples but note that a visualisation as in Figure~\ref{fig:atlasexamples} could be produced on a per-case basis in a deployed system aiding interpretability and visual means for manual validation by human experts.

\subsection*{(A) Initial Validation Study}

\begin{figure*}[tbh!]
    \includegraphics[width=0.95\textwidth]{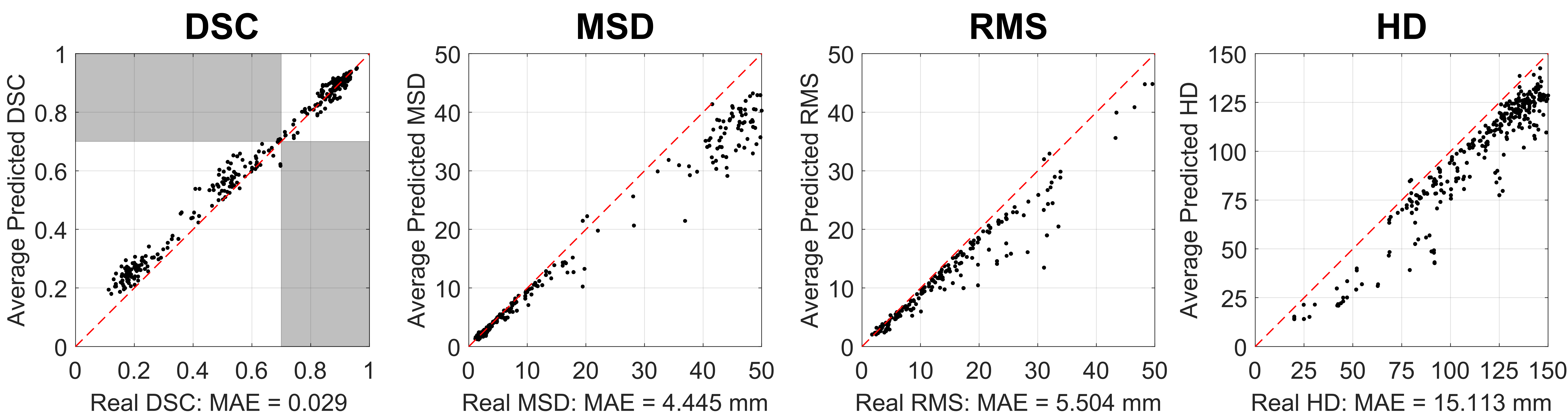}
	\caption{\csentence{RCA Validation on 400 cardiac MRI.}
		400 cardiac MRI segmentations were generated with a Random Forest classifier. 500 trees and depths in the range $[5, \ 40]$ were used to simulate various degrees of segmentation quality. RCA with single-atlas classifier was used to predict the Dice Similarity Coefficient (DSC), mean surface distance (MSD), root mean-squared surface distance (RMS) and Hausdorff distance (HD). Ground truth for the scans is known so real metrics are also calculated. All calculations on the whole-heart binary classification task. We report low mean absolute error (MAE) for all metrics and 99\% binary classification accuracy (TPR = 0.98, FPR = 0.00) with a DSC threshold of 0.70. High accuracy for individual segmentation classes. Absolute error for each image is shown for each metric. We note increasing error with decreasing quality of segmentation based on the real metric score.}
		\label{fig:RCAatlas}
\end{figure*}

A summary of the results is shown in Table~\ref{tab:RCAatlas}. We observe low mean absolute error (MAE) across all evaluation metrics and all class labels. The scatter plots in Figure~\ref{fig:RCAatlas} on real and predicted scores illustrate the very good performance of RCA in predicting segmentation quality scores. We also find that from the 400 test segmentations, RCA is able to classify `good' ($\mathrm{DSC} \in \left[ 0.7 \ 1.0\right]$) and `poor' ($\mathrm{DSC} \in [0.1 \ 0.7)$) segmentations with an accuracy of 99\%. From 171 poor segmentations at this threshold, 166 could be correctly identified by RCA, i.e. 97.1\%. 100\% of good-quality segmentations were correctly labelled. Additionally, we find binary classification accuracy of 95\% when applying a threshold of 2.0 mm on the MSD. From 365 poor segmentations at this threshold, 348 could be correctly identified by RCA, i.e. 95.3\%. Similarly, 31 from 35 (88.6\%) good-quality segmentations were correctly labelled. For all evaluation metrics, there is a strong, positive linear relationship between predicted and real values with $r \in \left[ 0.95 \ 0.99 \right]$ and $p < 0.0001$. Further analysis of our data shows increasing absolute error in each metric as the real score gets worse, e.g. the error for MSD increases with increasing surface distance. This correlates larger MAE with lower segmentation quality. In addition, when we consider only those segmentations where the real metric is 30 or less, the MAE drop significantly to 0.65, 1.71 and 6.78 mm for MSD, RMS and HD respectively. We are not concerned with greater errors for poor segmentations as they are still likely to be identified by RCA as having failed.

\begin{table}[tbh!]
\caption{\textbf{Initial Reverse Classification Accuracy Validation on 400 Random Forest Segmentations.} Classes are LV Cavity (LVC), LV Myocardium (LVM), RV Cavity (RVC), An average over the classes (Av.) and a binary segmentation of the whole heart (WH).  First row for each class shows the binary classification accuracy for `poor' and `good' segmentations in the Dice Similarity Coefficient (DSC) ranges $\left[ 0.0 \ 0.7 \right)$ and $\left[ 0.7 \ 1.0 \right]$ respectively. Second row for each class shows the binary classification accuracy for `poor' and `good' segmentations in the Mean Surface Distance (MSD) ranges $\left[ > 2.0 \mathrm{mm} \right]$ and $\left[ 0.0\mathrm{mm} \ 2.0 \mathrm{mm} \right]$ respectively. True-positive and false-positive rates are also shown. We report mean absolute errors (MAE) on the predictions of DSC and additional surface-distance metrics: root-mean-squared surface distance (RMS) and Hausdorff distance (HD).}
	\label{tab:RCAatlas}
      \begin{tabular}{rccc|cccc}
        																														\hline
        \multirow{3}{*}{Class}  		& \multirow{3}{*}{Acc.}& \multirow{3}{*}{TPR}& \multirow{3}{*}{FPR}& \multicolumn{4}{c}{MAE}\\	\cline{5-8}				
				           				& 			&	&			& DSC		& MSD			& RMS  			& HD			\\
				           				&			&	&			&			& \textit{mm}	& \textit{mm}	& \textit{mm}	\\	\hline
				        LVC 		& 0.973				&0.977	& 0.036		& \multirow{2}{*}{0.020} 		& \multirow{2}{*}{4.104}			& \multirow{2}{*}{5.593} 		& \multirow{2}{*}{14.15}				\\
& 0.980 & 0.975 & 0.019 & & & & \\[0.1cm]
				        LVM 		& 0.815				&0.947	& 0.215		& \multirow{2}{*}{0.044} 		& \multirow{2}{*}{3.756}			& \multirow{2}{*}{4.741}  		& \multirow{2}{*}{13.08}				\\
& 0.990 & 0.987 & 0.008 & & & & \\[0.1cm]
				        RVC 		& 0.985				&0.923	& 0.012		& \multirow{2}{*}{0.030}  		& \multirow{2}{*}{4.104}			& \multirow{2}{*}{5.022}   		& \multirow{2}{*}{16.63}				\\
& 0.943 & 0.914 & 0.047 & & & & \\[0.1cm]
                        Av.			& 0.924				&0.949	& 0.089		& \multirow{2}{*}{0.031}			& \multirow{2}{*}{3.988}			& \multirow{2}{*}{5.119}			& \multirow{2}{*}{14.62}				\\
& 0.971 & 0.959 & 0.025 & & & & \\[0.1cm]
				        WH			&\textbf{0.988}	&\textbf{0.979}	&\textbf{0.000}		&\multirow{2}{*}{\textbf{0.029}}& \multirow{2}{*}{4.445}			& \multirow{2}{*}{5.504}			& \multirow{2}{*}{15.11}				\\
& \textbf{0.948} & \textbf{0.886} & \textbf{0.047} & & & &\\	\hline
      \end{tabular}
\end{table}

\subsection*{(B) Large-scale Validation with Manual GT on UKBB}

\begin{figure*}[tbh!]
    \includegraphics[width=0.95\textwidth]{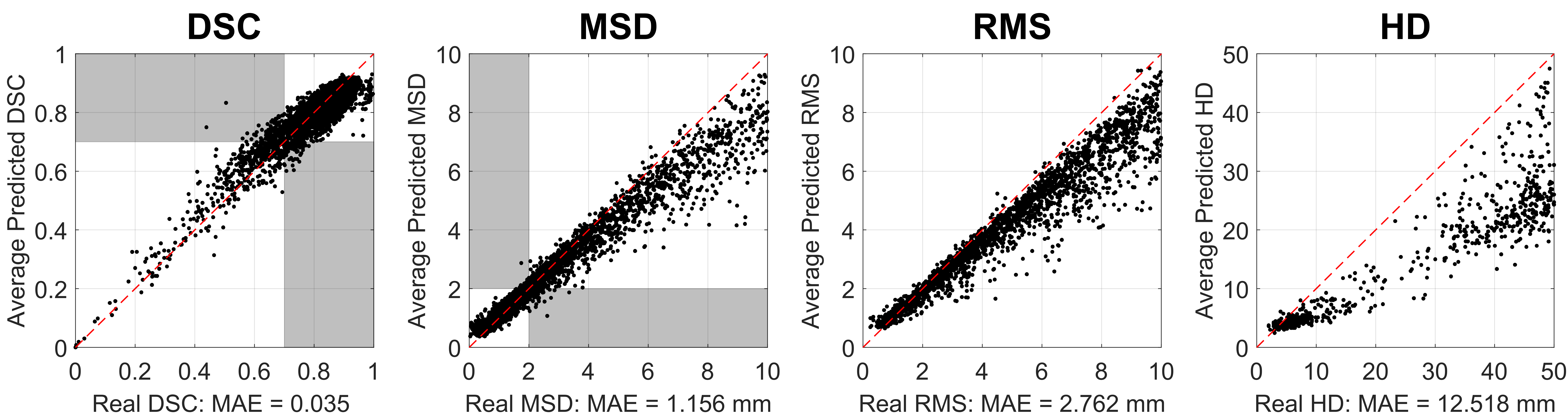}
	\caption{\csentence{Validation on 4,805 Random Forest segmentations of UKBB Imaging Study with Ground Truth.}
		4,805 cardiac MRI were segmented with a Random Forest classifier. 500 trees and depths in the range $[5 \ 40]$ were used to simulate various degrees of segmentation quality. Manual contours were available through Biobank Application 2964. RCA with single-atlas classifier was used to predict the Dice Similarity Coefficient (DSC), mean surface distance (MSD), root mean-squared surface distance (RMS) and Hausdorff distance (HD). All calculations on the whole-heart binary classification task. We report low mean absolute error (MAE) for all metrics and 95\% binary classification accuracy (TPR = 0.97 and FPR = 0.15) with a DSC threshold of 0.70. High accuracy for individual segmentation classes.}
			\label{fig:RCA5k}
\end{figure*}

Results for the RF segmentations are shown in Table~\ref{tab:RCA5k}. We report 95\% binary classification accuracy with a DSC threshold of 0.7 and low MAE on the DSC. From 589 poor segmentations at this threshold, 443 could be correctly identified by RCA, i.e. 75.2\%. Similarly, 4139 from 4216 (98.2\%) good-quality segmentations were correctly labelled. Additionally, we find binary classification accuracy of 98\% when applying a threshold of 2.0 mm on the MSD. From 2497 poor segmentations at this threshold, 2429 could be correctly identified by RCA, i.e. 97.3\%. Similarly, 2270 from 2308 (98.3\%) good-quality segmentations were correctly labelled. The true positive rates (TPR) are high across the classes, this shows RCA is able to correctly and consistently identify `good' quality segmentations. MSD-based false-positive rates (FPR) are shown to be lower than those based on DSC, this would indicate that MSD is more discriminative for `poor' quality segmentations and does not misclassify them so much as DSC. We identify only two instances where RCA predictions do not conform to the overall trend and predict much higher than the real DSC. On inspection, we find that the GT of these segmentations were missing mid-slices causing the real DSC to drop. These points can be seen in the upper-left-hand quadrant on Figure~\ref{fig:RCA5k}. The figure also shows that, over all metrics, there is high correlation between predicted and real quality metrics. This is very much comparable to the results from our initial validation study (A) in Figure~\ref{fig:RCAatlas}. The strong relationship between the predicted quality metrics from RCA and the equivalent scores calculated with respect to the manual segmentations demonstrates concretely that RCA is capable of correctly identifying, on a case-by-case basis, segmentations of poor quality in large-scale imaging studies.

\begin{figure*}[tbh!]
    \includegraphics[width=0.95\textwidth]{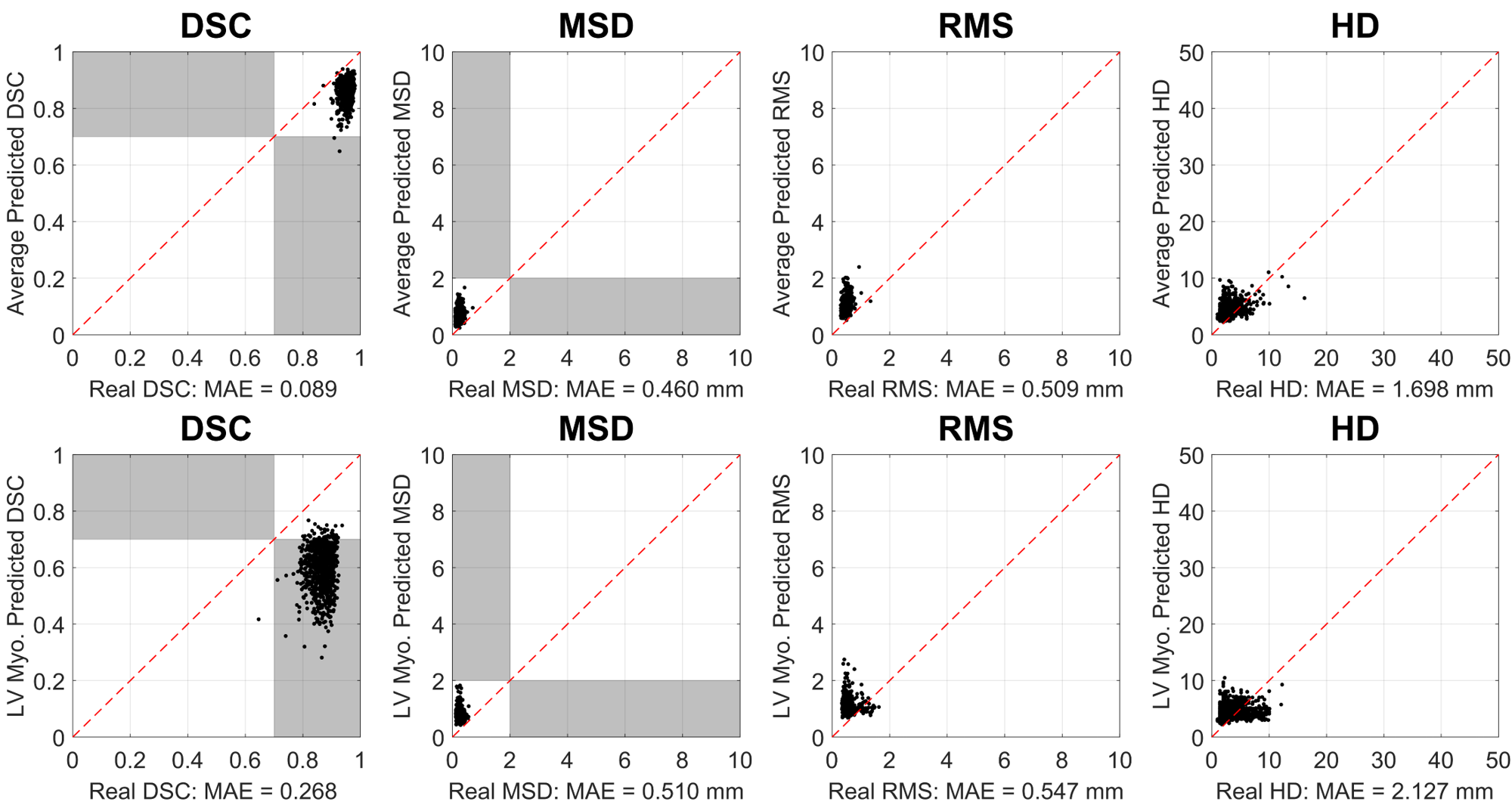}
	\caption{\csentence{Extensive Reverse Classification Accuracy Validation on 900 UKBB Segmentations.} Convolutional neural network (CNN) segmentation as in Bai et al. \cite{Bai2017}. Manual contours were available through Biobank Application 2964. RCA with single-atlas classifier was used to predict the Dice Similarity Coefficient (DSC), mean surface distance (MSD), root mean-squared surface distance (RMS) and Hausdorff distance (HD). All calculations for the binary quality classification task on (top) 'Whole Heart' average and (bottom) Left Ventricular Myocardium. We report low mean absolute error (MAE) for all metrics and 99.8\% binary classification accuracy (TPR = 1.00 and FPR = 0.00) with a DSC threshold of 0.70.}
			\label{fig:RCAcnn}
\end{figure*}

On the CNN segmentations, we report 99.8\% accuracy in binary classification for the whole-heart class. With a DSC threshold set at 0.7, RCA correctly identified 898 from 900 good-quality segmentations with 2 false-negatives. A visualization of this can be seen in the top panel of Figure~\ref{fig:RCAcnn} where the predicted and real DSC can be seen clustered in the high-quality corner of each metric's plot (upper-right for DSC and lower-left for surface-distance metrics). This reflects the high quality segmentations of the deep learning approach which have been correctly identified as such using RCA. Table~\ref{tab:RCA5kcnn} shows the detailed statistics for this experiment.

We note that the individual class accuracy for the LV myocardium is lower in the CNN case when DSC is used at the quality metric. We show the results for this class in the bottom panel of Figure~\ref{fig:RCAcnn}. Segmentors can have difficulty with this class due to its more complex shape. From the plotted points we see all cases fall into a similar cluster to the average WH case, but the RCA score under-predicts the real DSC. This exemplifies a task-specific setting for how RCA would be used in practice. In this case one cannot rely only on DSC to predict the quality of the segmentation, so MSD could provide a more appropriate quality prediction.

\begin{table}[tb!]
\caption{\textbf{Analysis of 4,800 Random Forest segmentations with available ground truth.} 4,800 RF segmentation at various depths $[5 \ 40]$ and 500 trees. Manual contours were available through Biobank Application 2964. Classes are LV Cavity (LVC), LV Myocardium (LVM), RV Cavity (RVC), an average over the classes (Av.) and a binary segmentation of the whole heart (WH). First row for each class shows the binary classification accuracy for `poor' and `good' segmentations in the Dice Similarity Coefficient (DSC) ranges $\left[ 0.0 \ 0.7 \right)$ and $\left[ 0.7 \ 1.0 \right]$ respectively. Second row for each class shows the binary classification accuracy for `poor' and `good' segmentations in the Mean Surface Distance (MSD) ranges $\left[ > 2.0 \mathrm{mm} \right]$ and $\left[ 0.0\mathrm{mm} \ 2.0 \mathrm{mm} \right]$ respectively. True-positive and false-positive rates are also shown. We report mean absolute errors (MAE) on the predictions of DSC and additional surface-distance metrics: root-mean-squared surface distance (RMS) and Hausdorff distance (HD).}
\label{tab:RCA5k}
      \begin{tabular}{rccc|cccc}
        																														\hline
        \multirow{3}{*}{Class}  		& \multirow{3}{*}{Acc.}& \multirow{3}{*}{TPR}& \multirow{3}{*}{FPR}& \multicolumn{4}{c}{MAE}\\	\cline{5-8}				
				           				& 			&	&			& DSC		& MSD			& RMS  			& HD			\\
				           				&			&	&			&			& \textit{mm}	& \textit{mm}	& \textit{mm}	\\	\hline
LVC 		& 0.968			& 0.997			& 0.330		& \multirow{2}{*}{0.042} 		& \multirow{2}{*}{0.906}			& \multirow{2}{*}{2.514} 		& \multirow{2}{*}{11.09}			\\
& 0.975 & 0.962 & 0.011 & & & & \\[0.1cm]
LVM 		& 0.454			& 0.956			& 0.571		& \multirow{2}{*}{0.125} 		& \multirow{2}{*}{0.963}			& \multirow{2}{*}{2.141}  		& \multirow{2}{*}{11.83}			\\
& 0.972 & 0.962 & 0.012 & & & & \\[0.1cm]
RVC 		& 0.868			& 0.957			& 0.352		& \multirow{2}{*}{0.057}			& \multirow{2}{*}{1.140}			& \multirow{2}{*}{2.790}  		& \multirow{2}{*}{15.23}			\\
& 0.969 & 0.977 & 0.040 & & & & \\[0.1cm]
Av.		& 0.763				& 0.970			& 0.418 	& \multirow{2}{*}{0.075}			& \multirow{2}{*}{1.003}			& \multirow{2}{*}{2.482}			& \multirow{2}{*}{12.72}			\\
& 0.972 & 0.967 & 0.032 & & & & \\[0.1cm]
WH		&\textbf{0.954}	&\textbf{0.966}	&\textbf{0.148}		& \multirow{2}{*}{\textbf{0.035}}& \multirow{2}{*}{1.156}			& \multirow{2}{*}{2.762}			& \multirow{2}{*}{12.52}			\\
& \textbf{0.978} & \textbf{0.984} & \textbf{0.027} & & & & \\\hline
      \end{tabular}
\end{table}

\
\begin{table}[tb!]
\caption{\textbf{Analysis of 900 CNN segmentations with available ground truth.} CNN segmentations as in Bai et al \cite{Bai2017}. Manual contours were available through Biobank Application 2964. Classes are LV Cavity (LVC), LV Myocardium (LVM), RV Cavity (RVC), an average over the classes (Av.) and a binary segmentation of the whole heart (WH). First row for each class shows the binary classification accuracy for `poor' and `good' segmentations in the Dice Similarity Coefficient (DSC) ranges $\left[ 0.0 \ 0.7 \right)$ and $\left[ 0.7 \ 1.0 \right]$ respectively. Second row for each class shows the binary classification accuracy for `poor' and `good' segmentations in the Mean Surface Distance (MSD) ranges $\left[ > 2.0 \mathrm{mm} \right]$ and $\left[ 0.0\mathrm{mm} \ 2.0 \mathrm{mm} \right]$ respectively. True-positive and false-positive rates are also shown. We report mean absolute errors (MAE) on the predictions of DSC and additional surface-distance metrics: root-mean-squared surface distance (RMS) and Hausdorff distance (HD).}
\label{tab:RCA5kcnn}
      \begin{tabular}{rccc|cccc}
        																														\hline
        \multirow{3}{*}{Class}  		& \multirow{3}{*}{Acc.}& \multirow{3}{*}{TPR}& \multirow{3}{*}{FPR}& \multicolumn{4}{c}{MAE}\\	\cline{5-8}				
				           				& 			&	&			& DSC		& MSD			& RMS  			& HD			\\
				           				&			&	&			&			& \textit{mm}	& \textit{mm}	& \textit{mm}	\\	\hline
LVC 		& 0.998			& 1.000			& 0.000		& \multirow{2}{*}{0.082} 		& \multirow{2}{*}{0.386}			& \multirow{2}{*}{0.442} 		& \multirow{2}{*}{1.344}			\\
& 1.000 & 1.000 & 0.000 & & & & \\[0.1cm]
LVM 		& 0.051			& 1.000			& 0.001		& \multirow{2}{*}{0.268} 		& \multirow{2}{*}{0.510}			& \multirow{2}{*}{0.547}  		& \multirow{2}{*}{2.127}			\\
& 1.000 & 1.000 & 0.000 & & & & \\[0.1cm]
RVC 		& 0.901			& 1.000			& 0.033		& \multirow{2}{*}{0.146}			& \multirow{2}{*}{0.588}			& \multirow{2}{*}{0.656}  		& \multirow{2}{*}{2.086}			\\
& 0.997 & 0.997 & 0.000 & & & & \\[0.1cm]
Av.		& 0.650				& 1.000			& 0.011 	& \multirow{2}{*}{0.165}			& \multirow{2}{*}{0.495}			& \multirow{2}{*}{0.548}			& \multirow{2}{*}{1.852}			\\
& 0.999 & 0.999 & 0.000 & & & & \\[0.1cm]
WH		&\textbf{0.998}	&\textbf{1.000}	&\textbf{0.000}		& \multirow{2}{*}{\textbf{0.089}}& \multirow{2}{*}{0.460}			& \multirow{2}{*}{0.509}			& \multirow{2}{*}{1.698}			\\
& \textbf{1.000} & \textbf{1.000} & \textbf{0.000} & & & & \\\hline
      \end{tabular}
\end{table}

\subsection*{(C) Quality Control on 7,250 UK Biobank Images}

\begin{figure*}[tbh!]
    \includegraphics[width=0.95\textwidth]{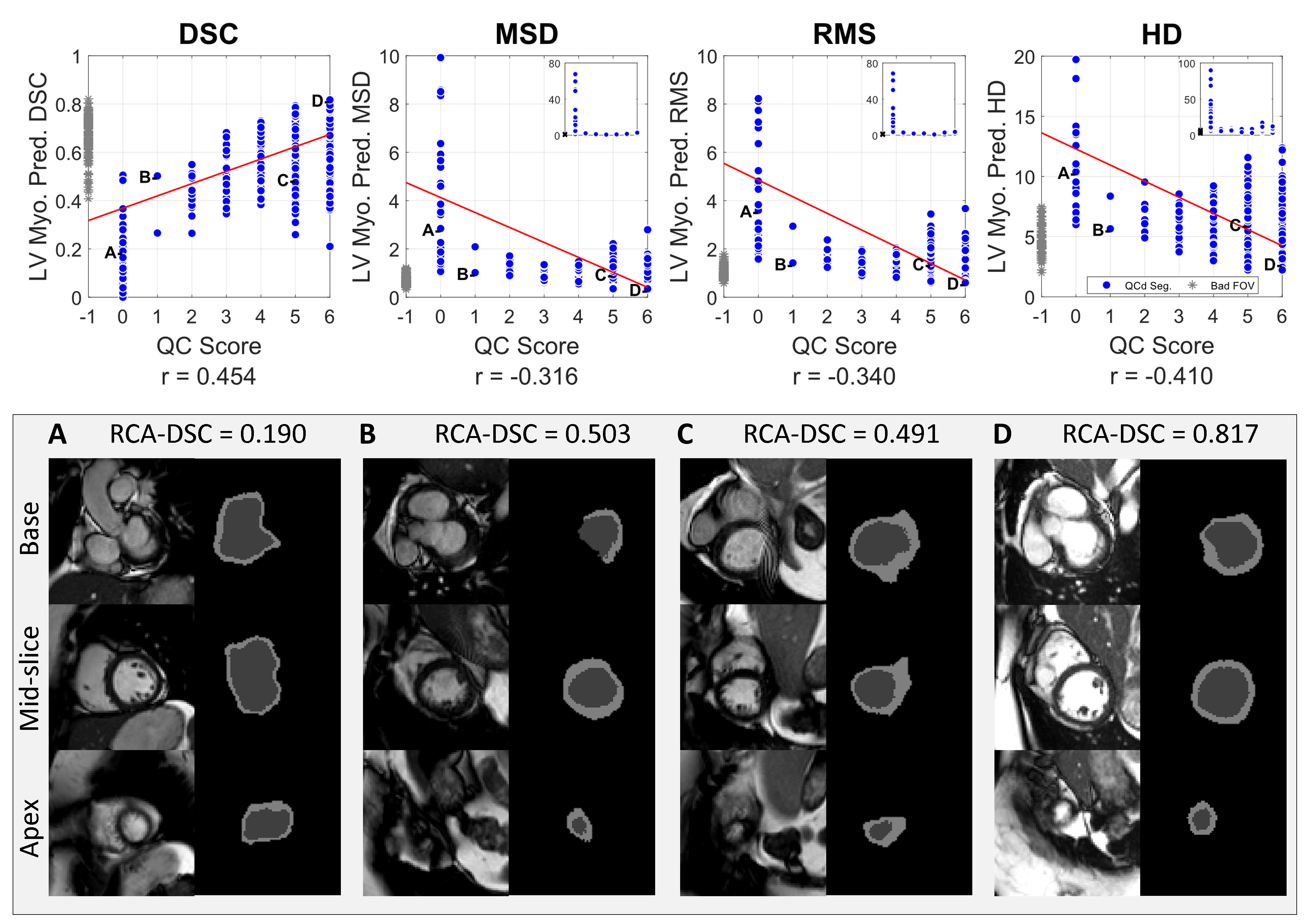}
	\caption{\csentence{RCA Application on 7,250 Cardiac MRI segmentations of UKBB Imaging Study.}
		7,250 cardiac MRI segmentations generated with a multi-atlas segmentation approach \cite{Bai2013}. Manual QC scores given in the range $[0 \ 6]$ (i.e. $[0 \ 2]$ for each of basal, mid and apical slices). RCA with single-atlas classifier was used to predict the Dice Similarity Coefficient (DSC), mean surface distance (MSD), root mean-squared surface distance (RMS) and Hausdorff distance (HD). All calculations on the LV Myocardium binary classification task. We show correlation in all metrics. Examples show: A) and B) agreement between low predicted DSC and low manual QC score, C) successful automated identification of poor segmentation with low predicted DSC despite high manual QC score and D) agreement between high predicted DSC and high manual QC score. Inserts in top row display extended range of y-axis.}
        \label{fig:RCA7kcombined}
\end{figure*}

Figure~\ref{fig:RCA7kcombined} shows the relationship between manual QC scores and the predicted DSC, MSD, RMS and HD obtained from RCA. Note, these predictions are for the LV myocardium and not the overall segmentation as this class was the focus of the manual QC procedure. Manual QC was not performed for the other classes.

Figure~ \ref{fig:RCA7kcombined} also shows a sample of segmentations with manual QC scores of 0, 1, 5 and 6 for the LV myocardium. With a score of 0, `A' must have a `poor' quality segmentation of LV myocardium at the basal, apical and mid slices. Example `B' shows relatively low surface-distance metrics and a low DSC, we see this visually as the boundary of the myocardium is in the expected region, but is incomplete in all slices. This segmentation has been given a score of 1 because the mid-slice is well segmented while the rest is not; which is correctly identified by RCA. In example `C', the segmentation of the LV myocardium is clearly not good with respect to the image, yet it has been given a manual QC score of 5. Again, RCA is able to pick up such outliers by predicting a lower DSC. The final example `D' displays an agreement between the high predicted DSC from RCA and the high manual QC score. These examples demonstrate RCA's ability to correctly identify both good and poor quality segmentations when performing assessments over an entire 3D segmentation. It also demonstrates the limitations of manual QC and the success of RCA in identifying segmentation failure on a per-case basis.

Creating a set of manual QC scores for over 7,200 images is a laborious task but it has provided worthwhile evidence for the utility of RCA on large-scale studies. It is clear, however, that the 3-layer inspection approach with a single-rater has limitations. First, inspecting all layers would be preferable, but it highly time-consuming and, second, with multiple raters, averaging or majority voting could be employed to reduce human error.

We should note that RCA is unlikely to mimic the exact visual manual QC process, and it should not, as it naturally provides a different, more comprehensive, assessment of segmentation quality. The manual QC is a rather crude assessment of segmentation quality, as such, we did not perform a direct, quantitative comparison using the visual QC categories but rather wanted to demonstrate that there is a general correlation between manual QC and the predicted RCA score.


\section*{Discussion}

We have shown that it is possible to assess the quality of individual cardiac MR segmentations at large-scale and in the absence of ground truth. Previous approaches have primarily focused on evaluating overall, average performance of segmentation methods or required large sets of pre-annotated data of good and bad quality segmentations for training a classifier. Our method is well suited for use in image-analysis pipelines and clinical workflows where the quality of segmentations should be assessed on a per-case basis. We have also shown that RCA can provide predictions on a per-class basis. Note that our manually labelled dataset did not include the RV Myocardium as a label and therefore has been omitted from our study.

The RCA validation process was carried out on 8-core Intel i7 3.6 GHz machines. The whole process for a single test segmentation - including 100 reference image registrations, warping 100 reference segmentations and metric evaluations - took on average 11 minutes, making it suitable for background processing in large-scale studies and clinical practice. However, this is a limitation as the runtime per case currently does not allow immediate feedback and prohibits applications with real-time constraints. For example, one could envision a process where cardiac MR scans are immediately segmented after acquisition, and feedback on the quality would be required while the patient is still in the scanner. For this, the computation time of RCA would need to be reduced possibly through an automatic selection of a subset of reference images. We report preliminary results for using a deep learning approach to speed up the process in \cite{Robinson2018}. With a real-time RCA framework, the method could be used to identify challenging cases for CNN-based segmentors where the RCA feedback could be used to improve the segmentation algorithm.

\begin{figure*}[h!]
    \includegraphics[width=0.45\textwidth]{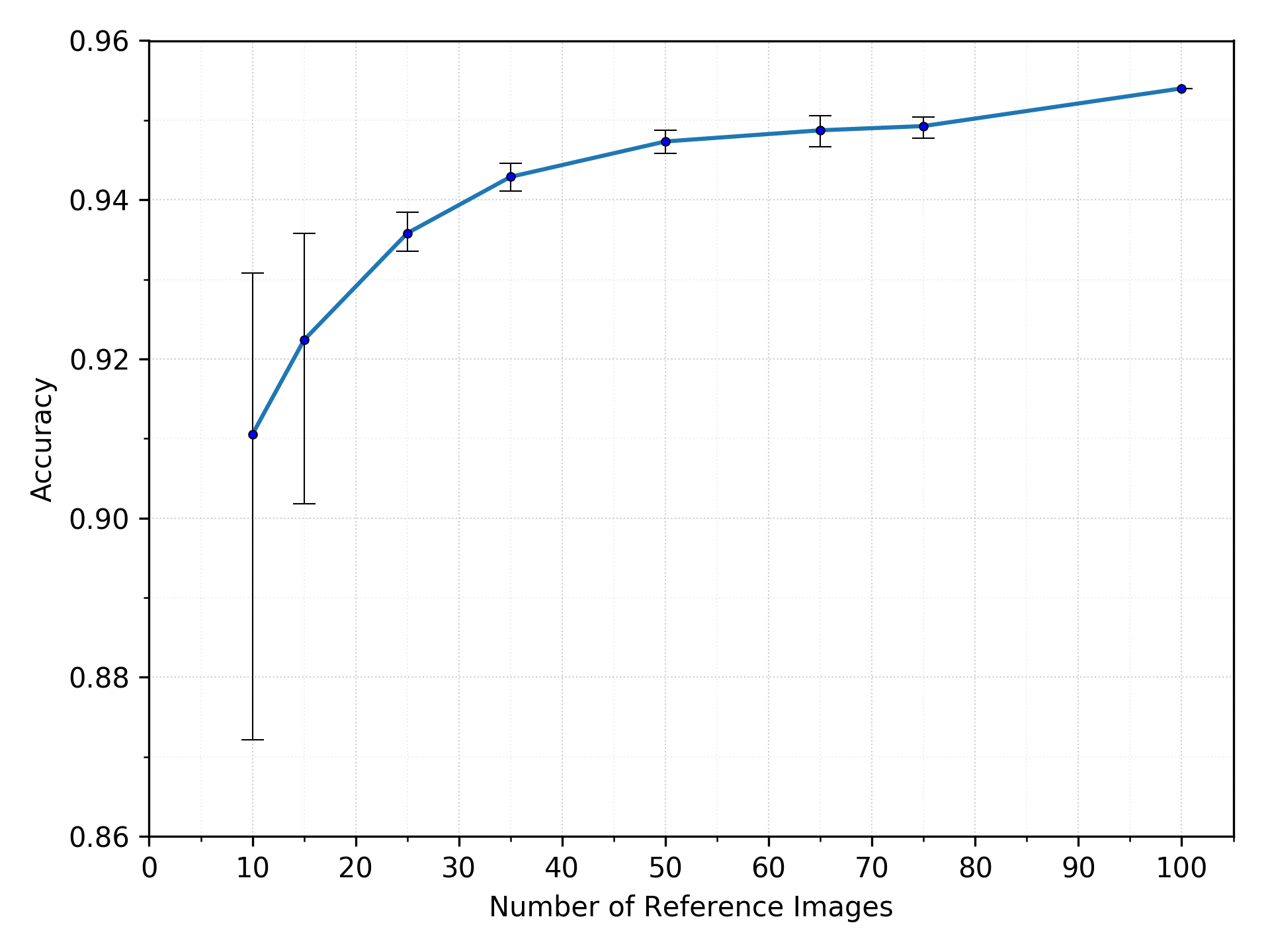}
	\caption{\rev{\csentence{Investigating the Effect of Reference Set Size on Prediction Accuracy.}
		4,805 automated segmentations from Experiment B were processed with Reverse Classification Accuracy (RCA) using differing numbers of reference images. Random subsets of 10, 15, 35, 50, 65 and 75 reference images were taken from the full set of 100 available reference images. Five random runs were performed to obtain error bars for each setting. Average prediction accuracy increases with increasing number of reference images and the variance between runs also decreases.}}
        \label{fig:refsexpt}
\end{figure*}

\rev{As noted earlier, using a subset of the reference set could help to optimize the run-time of RCA predictions. To better understand the effect of reference set size on prediction accuracy, we have performed an empirical evaluation using the data from Experiment B. We took the 4,805 automated segmentations and their manual GT and performed RCA using randomly selected subsets of the 100 image-segmentation pairs from the full reference set. Five different randomly selected sets of sizes 10, 15, 25, 35, 50, 65 and 75 were created and used for obtaining RCA predictions on the 4,805 images. Figure~\ref{fig:refsexpt} shows the mean accuracy computed across the 5 runs for each reference set size. Error bars indicate the highest and lowest accuracy achieved across the five runs. Accuracy is computed using the same DSC threshold of 0.7 as used in Experiment B. The figure shows that the mean accuracy increases with increasing number of reference images. The error bars in Figure~\ref{fig:refsexpt} show a decrease in size with increasing size of the reference set. As the reference set grows in size, a greater variability in the images is captured that allows the RCA process to become more accurate. Noteworthy, even with small reference sets of about 20 images high accuracy of more then 90$\%$ is obtained.}

Although RCA can give a good indication of the real DSC score for an individual segmentation, an accurate one-to-one mapping between the predicted and real DSC has not been achieved. However, we have shown that the method will confidently differentiate between `good' and `poor' quality segmentations based on an application specific threshold. The threshold could be chosen depending on the application's requirements for what qualifies as a `good' segmentation. Failed segmentations could be re-segmented with different parameters, regenerated with alternative methods, discarded from further analyses or, more likely, sent to a user for manual inspection. Additionally, whilst RCA has been shown to be robust to cardiovascular topology it would need to be re-evaluated for use in other anatomical regions.

\section*{Conclusion}

Reverse Classification Accuracy had previously been shown to effectively predict the quality of whole-body multi-organ segmentations. We have successfully validated the RCA framework on 3D cardiac MR, demonstrating the robustness of the methodology to different anatomy. RCA has been successful in identifying poor-quality image segmentations with measurements of DSC, MSD, RMS and HD and has shown excellent MAE against all metrics. RCA has also been successful in producing a comparable outcome to a manual quality control procedure on a large database of 7,250 images from the UKBB. We have shown further success in accurately predicting quality metrics on 4,805 segmentations from Petersen et al., for which manual segmentations were available for evaluation. Predicting segmentation accuracy in the absence of ground truth is a step towards fully automated QC in image analysis pipelines.

Our contributions to the field are three-fold: 1) a thorough validation of RCA for the application of cardiac MR segmentation QC. Our results indicate highly accurate predictions of segmentation quality across various metrics; 2) a feasibility study of using RCA for automatic QC in large-scale studies. RCA predictions correlate with a set of manual QC scores and enable outlier detection in a large set of 7,250 cases, and 3) a large-scale validation on 4,800 cardiac MR images from the UKBB. Furthermore, we have done this without the need for a large, labelled dataset and we can predict segmentation quality on a per-case basis.


\begin{backmatter}

\section*{Abbreviations}
CMR: cardiovascular magnetic resonance; LV: left ventricle; RV: right ventricle; ED: end-diastole; RCA: reverse classification accuracy; GT: ground truth; DSC: Dice similarity coefficient; MSD: mean surface distance; RMS: root-mean-squared  surface distance; HD: Hausdorff distance; MAE: mean absolute error; RF: random forest; CNN: convolutional neural network; UKBB: UK Biobank; 3D: 3-dimensional; MR: magnetic resonance; CMR: cardiac magnetic resonance; QC: quality control; GWAS: genome-wide association study; FOV: field of view; TPR: true-positive rate; FPR: false-positive rate; AF: atlas forest;

\section*{Ethics approval and consent to participate}
The UKBB has approval from the North West Research Ethics Committee (REC reference: 11/NW/0382). All participants have given written informed consent.


\section*{Availability of data and materials}
The imaging data and manual annotations were provided by the UKBB Resource under Application Number 2946. Part (B) was conducted with data obtained through Application Number 18545. Researchers can apply to use the UKBB data resource for health-related research in the public interest. Python code for performing RCA as implemented in this work is in the public repository: https://github.com/mlnotebook/RCA. Example sets of reference and test-images are provided as links in the repository.

\section*{Consent for Publication}
Not applicable

\section*{Competing interests}
Steffen E. Petersen provides consultancy to Circle Cardiovascular Imaging Inc. (Calgary, Alberta, Canada). Ben Glocker receives research funding from HeartFlow Inc. (Redwood City, CA, USA).

\section*{Author's contributions}
RR and BG conceived and designed the study; RR performed implementations, data analysis and write the manuscript. VV developed the original RCA framework. WB provided the automated segmentations that were given QC scores by HS. OO provided landmarks used in the original version of this work. SN, SKP and SEP are overall consortium leads for UKBB access application 2964 and responsible for the conceptualisation of creating a CMR segmentation reference standard. Data curation for application 2964 by SEP, NA, AML and VC and manual contours of 5,000 CMR produced by MMS, NA, JMP, FZ, KF, EL, VC and YJK. PM and DR provided advice in the stage of model development and clinical applications.

\section*{Funding}
RR is funded by both the King’s College London \& Imperial College London EPSRC Centre for Doctoral Training in Medical Imaging (EP/L015226/1) and GlaxoSmithKline; VV by Indonesia Endowment for Education (LPDP) Indonesian Presidential PhD Scholarship and HS by Research Fellowship from Uehara Memorial Foundation. This work was also supported by the following institutions: KF is supported by The Medical College of Saint Bartholomew’s Hospital Trust, an independent registered charity that promotes and advances medical and dental education and research at Barts and The London School of Medicine and Dentistry. AL and SEP acknowledge support from the NIHR Barts Biomedical Research Centre and from the ``SmartHeart'' EPSRC program grant (EP/P001009/ 1). SN and SKP are supported by the Oxford NIHR Biomedical Research Centre and the Oxford British Heart Foundation Centre of Research Excellence. This project was enabled through access to the MRC eMedLab Medical Bioinformatics infrastructure, supported by the Medical Research Council (grant number MR/L016311/1). NA is supported by a Wellcome Trust Research Training Fellowship (203553/Z/Z). The authors SEP, SN and SKP acknowledge the British Heart Foundation (BHF) for funding the manual analysis to create a cardiovascular magnetic resonance imaging reference standard for the UKBB imaging resource in 5000 CMR scans (PG/14/89/31194). PMM gratefully acknowledges support from the Edmond J. Safra Foundation and Lily Safra, the Imperial College Healthcare Trust Biomedical Research Centre, the EPSRC Centre for Mathematics in Precision Healthcare, the UK Dementia Research Institute and the MRC. BG received funding from the European Research Council (ERC) under the European Union's Horizon 2020 research and innovation programme (grant agreement No 757173, project MIRA, ERC-2017-STG).

\section*{Acknowledgements}
This work was carried out under UKBB Applications 18545 and 2964. The authors wish to thank all UKBB participants and staff. 


\bibliographystyle{bmc-mathphys} 
\bibliography{jcmr-rca-robinson}      


\begin{thebibliography}{19}
\ifx \bisbn   \undefined \def \bisbn  #1{ISBN #1}\fi
\ifx \binits  \undefined \def \binits#1{#1}\fi
\ifx \bauthor  \undefined \def \bauthor#1{#1}\fi
\ifx \batitle  \undefined \def \batitle#1{#1}\fi
\ifx \bjtitle  \undefined \def \bjtitle#1{#1}\fi
\ifx \bvolume  \undefined \def \bvolume#1{\textbf{#1}}\fi
\ifx \byear  \undefined \def \byear#1{#1}\fi
\ifx \bissue  \undefined \def \bissue#1{#1}\fi
\ifx \bfpage  \undefined \def \bfpage#1{#1}\fi
\ifx \blpage  \undefined \def \blpage #1{#1}\fi
\ifx \burl  \undefined \def \burl#1{\textsf{#1}}\fi
\ifx \doiurl  \undefined \def \doiurl#1{\textsf{#1}}\fi
\ifx \betal  \undefined \def \betal{\textit{et al.}}\fi
\ifx \binstitute  \undefined \def \binstitute#1{#1}\fi
\ifx \binstitutionaled  \undefined \def \binstitutionaled#1{#1}\fi
\ifx \bctitle  \undefined \def \bctitle#1{#1}\fi
\ifx \beditor  \undefined \def \beditor#1{#1}\fi
\ifx \bpublisher  \undefined \def \bpublisher#1{#1}\fi
\ifx \bbtitle  \undefined \def \bbtitle#1{#1}\fi
\ifx \bedition  \undefined \def \bedition#1{#1}\fi
\ifx \bseriesno  \undefined \def \bseriesno#1{#1}\fi
\ifx \blocation  \undefined \def \blocation#1{#1}\fi
\ifx \bsertitle  \undefined \def \bsertitle#1{#1}\fi
\ifx \bsnm \undefined \def \bsnm#1{#1}\fi
\ifx \bsuffix \undefined \def \bsuffix#1{#1}\fi
\ifx \bparticle \undefined \def \bparticle#1{#1}\fi
\ifx \barticle \undefined \def \barticle#1{#1}\fi
\ifx \bconfdate \undefined \def \bconfdate #1{#1}\fi
\ifx \botherref \undefined \def \botherref #1{#1}\fi
\ifx \url \undefined \def \url#1{\textsf{#1}}\fi
\ifx \bchapter \undefined \def \bchapter#1{#1}\fi
\ifx \bbook \undefined \def \bbook#1{#1}\fi
\ifx \bcomment \undefined \def \bcomment#1{#1}\fi
\ifx \oauthor \undefined \def \oauthor#1{#1}\fi
\ifx \citeauthoryear \undefined \def \citeauthoryear#1{#1}\fi
\ifx \endbibitem  \undefined \def \endbibitem {}\fi
\ifx \bconflocation  \undefined \def \bconflocation#1{#1}\fi
\ifx \arxivurl  \undefined \def \arxivurl#1{\textsf{#1}}\fi
\csname PreBibitemsHook\endcsname

\bibitem{Sudlow2015}
\begin{barticle}
\bauthor{\bsnm{Sudlow}, \binits{C.}},
\bauthor{\bsnm{Gallacher}, \binits{J.}},
\bauthor{\bsnm{Allen}, \binits{N.}},
\bauthor{\bsnm{Beral}, \binits{V.}},
\bauthor{\bsnm{Burton}, \binits{P.}},
\bauthor{\bsnm{Danesh}, \binits{J.}},
\bauthor{\bsnm{Downey}, \binits{P.}},
\bauthor{\bsnm{Elliott}, \binits{P.}},
\bauthor{\bsnm{Green}, \binits{J.}},
\bauthor{\bsnm{Landray}, \binits{M.}},
\bauthor{\bsnm{Liu}, \binits{B.}},
\bauthor{\bsnm{Matthews}, \binits{P.}},
\bauthor{\bsnm{Ong}, \binits{G.}},
\bauthor{\bsnm{Pell}, \binits{J.}},
\bauthor{\bsnm{Silman}, \binits{A.}},
\bauthor{\bsnm{Young}, \binits{A.}},
\bauthor{\bsnm{Sprosen}, \binits{T.}},
\bauthor{\bsnm{Peakman}, \binits{T.}},
\bauthor{\bsnm{Collins}, \binits{R.}}:
\batitle{{UK Biobank: An Open Access Resource for Identifying the Causes of a
  Wide Range of Complex Diseases of Middle and Old Age}}.
\bjtitle{PLoS Medicine}
\bvolume{12}(\bissue{3}),
\bfpage{1}--\blpage{10}
(\byear{2015}).
doi:\doiurl{10.1371/journal.pmed.1001779}
\end{barticle}
\endbibitem

\bibitem{Shariff2010}
\begin{barticle}
\bauthor{\bsnm{Shariff}, \binits{A.}},
\bauthor{\bsnm{Kangas}, \binits{J.}},
\bauthor{\bsnm{Coelho}, \binits{L.P.}},
\bauthor{\bsnm{Quinn}, \binits{S.}},
\bauthor{\bsnm{Murphy}, \binits{R.F.}}:
\batitle{{Automated Image Analysis for High-Content Screening and Analysis}}.
\bjtitle{Journal of Biomolecular Screening}
\bvolume{15}(\bissue{7}),
\bfpage{726}--\blpage{734}
(\byear{2010}).
doi:\doiurl{10.1177/1087057110370894}
\end{barticle}
\endbibitem

\bibitem{DeBruijne2016}
\begin{barticle}
\bauthor{\bparticle{de} \bsnm{Bruijne}, \binits{M.}}:
\batitle{{Machine learning approaches in medical image analysis: From detection
  to diagnosis.}}
\bjtitle{Medical Image Analysis}
\bvolume{33},
\bfpage{94}--\blpage{97}
(\byear{2016}).
doi:\doiurl{10.1016/j.media.2016.06.032}
\end{barticle}
\endbibitem

\bibitem{Bai2017}
\begin{botherref}
\oauthor{\bsnm{Bai}, \binits{W.}},
\oauthor{\bsnm{Sinclair}, \binits{M.}},
\oauthor{\bsnm{Tarroni}, \binits{G.}},
\oauthor{\bsnm{Oktay}, \binits{O.}},
\oauthor{\bsnm{Rajchl}, \binits{M.}},
\oauthor{\bsnm{Vaillant}, \binits{G.}},
\oauthor{\bsnm{Lee}, \binits{A.M.}},
\oauthor{\bsnm{Aung}, \binits{N.}},
\oauthor{\bsnm{Lukaschuk}, \binits{E.}},
\oauthor{\bsnm{Sanghvi}, \binits{M.M.}},
\oauthor{\bsnm{Zemrak}, \binits{F.}},
\oauthor{\bsnm{Fung}, \binits{K.}},
\oauthor{\bsnm{Paiva}, \binits{J.M.}},
\oauthor{\bsnm{Carapella}, \binits{V.}},
\oauthor{\bsnm{Kim}, \binits{Y.J.}},
\oauthor{\bsnm{Suzuki}, \binits{H.}},
\oauthor{\bsnm{Kainz}, \binits{B.}},
\oauthor{\bsnm{Matthews}, \binits{P.M.}},
\oauthor{\bsnm{Petersen}, \binits{S.E.}},
\oauthor{\bsnm{Piechnik}, \binits{S.K.}},
\oauthor{\bsnm{Neubauer}, \binits{S.}},
\oauthor{\bsnm{Glocker}, \binits{B.}},
\oauthor{\bsnm{Rueckert}, \binits{D.}}:
Human-level cmr image analysis with deep fully convolutional networks.
\arxivurl{1710.09289v3}
\end{botherref}
\endbibitem

\bibitem{Crum2006}
\begin{barticle}
\bauthor{\bsnm{Crum}, \binits{W.R.}},
\bauthor{\bsnm{Camara}, \binits{O.}},
\bauthor{\bsnm{Hill}, \binits{D.L.G.}}:
\batitle{{Generalized overlap measures for evaluation and validation in medical
  image analysis}}.
\bjtitle{IEEE Transactions on Medical Imaging}
\bvolume{25}(\bissue{11}),
\bfpage{1451}--\blpage{1461}
(\byear{2006}).
doi:\doiurl{10.1109/TMI.2006.880587}
\end{barticle}
\endbibitem

\bibitem{Taha2015}
\begin{barticle}
\bauthor{\bsnm{Taha}, \binits{A.A.}},
\bauthor{\bsnm{Hanbury}, \binits{A.}}:
\batitle{{Metrics for evaluating 3D medical image segmentation: analysis,
  selection, and tool.}}
\bjtitle{BMC medical imaging}
\bvolume{15},
\bfpage{29}
(\byear{2015}).
doi:\doiurl{10.1186/s12880-015-0068-x}
\end{barticle}
\endbibitem

\bibitem{Carapella2016a}
\begin{bchapter}
\bauthor{\bsnm{Carapella}, \binits{V.}},
\bauthor{\bsnm{Jim{\'{e}}nez-Ruiz}, \binits{E.}},
\bauthor{\bsnm{Lukaschuk}, \binits{E.}},
\bauthor{\bsnm{Aung}, \binits{N.}},
\bauthor{\bsnm{Fung}, \binits{K.}},
\bauthor{\bsnm{Paiva}, \binits{J.}},
\bauthor{\bsnm{Sanghvi}, \binits{M.}},
\bauthor{\bsnm{Neubauer}, \binits{S.}},
\bauthor{\bsnm{Petersen}, \binits{S.}},
\bauthor{\bsnm{Horrocks}, \binits{I.}},
\bauthor{\bsnm{Piechnik}, \binits{S.}}:
\bctitle{{Towards the Semantic Enrichment of Free-Text Annotation of Image
  Quality Assessment for UK Biobank Cardiac Cine MRI Scans}}.
In: \bbtitle{MICCAI Workshop on Large-scale Annotation of Biomedical Data and
  Expert Label Synthesis (LABELS)},
pp. \bfpage{238}--\blpage{248}.
\bpublisher{Springer},
\blocation{{Cham}}
(\byear{2016}).
doi:\doiurl{10.1007/978-3-319-46976-8\_25}
\end{bchapter}
\endbibitem

\bibitem{Zhang_2016}
\begin{bchapter}
\bauthor{\bsnm{Zhang}, \binits{L.}},
\bauthor{\bsnm{Gooya}, \binits{A.}},
\bauthor{\bsnm{Dong}, \binits{B.}},
\bauthor{\bsnm{Hua}, \binits{R.}},
\bauthor{\bsnm{Petersen}, \binits{S.E.}},
\bauthor{\bsnm{Medrano-Gracia}, \binits{P.}},
\bauthor{\bsnm{Frangi}, \binits{A.F.}}:
\bctitle{{Automated Quality Assessment of Cardiac MR Images Using Convolutional
  Neural Networks}}.
In: \beditor{\bsnm{Tsaftaris}, \binits{S.A.}},
\beditor{\bsnm{Gooya}, \binits{A.}},
\beditor{\bsnm{Frangi}, \binits{A.F.}},
\beditor{\bsnm{Prince}, \binits{J.L.}} (eds.)
\bbtitle{Medical Image Computing and Computer-Assisted Intervention – SASHIMI
  2016}.
\bsertitle{Lecture Notes in Computer Science},
vol. \bseriesno{9968},
pp. \bfpage{138}--\blpage{145}.
\bpublisher{Springer},
\blocation{{Cham}}
(\byear{2016}).
doi:\doiurl{10.1007/978-3-319-46630-9_14}
\end{bchapter}
\endbibitem

\bibitem{Zhong2010}
\begin{bchapter}
\bauthor{\bsnm{Zhong}, \binits{E.}},
\bauthor{\bsnm{Fan}, \binits{W.}},
\bauthor{\bsnm{Yang}, \binits{Q.}},
\bauthor{\bsnm{Verscheure}, \binits{O.}},
\bauthor{\bsnm{Ren}, \binits{J.}}:
\bctitle{{Cross Validation Framework to Choose amongst Models and Datasets for
  Transfer Learning}}.
In: \bbtitle{Lecture Notes in Computer Science (including Subseries Lecture
  Notes in Artificial Intelligence and Lecture Notes in Bioinformatics)}
vol. \bseriesno{6323 LNAI},
pp. \bfpage{547}--\blpage{562}.
\bpublisher{Springer}, \blocation{???}
(\byear{2010}).
doi:\doiurl{10.1007/978-3-642-15939-8_35}.
\burl{http://link.springer.com/10.1007/978-3-642-15939-8{\_}35}
\end{bchapter}
\endbibitem

\bibitem{Fan2006}
\begin{bchapter}
\bauthor{\bsnm{Fan}, \binits{W.}},
\bauthor{\bsnm{Davidson}, \binits{I.}}:
\bctitle{{Reverse testing}}.
In: \bbtitle{Proceedings of the 12th ACM SIGKDD International Conference on
  Knowledge Discovery and Data Mining - KDD '06},
p. \bfpage{147}.
\bpublisher{ACM Press},
\blocation{New York, New York, USA}
(\byear{2006}).
doi:\doiurl{10.1145/1150402.1150422}
\end{bchapter}
\endbibitem

\bibitem{Valindriaa}
\begin{botherref}
\oauthor{\bsnm{Valindria}, \binits{V.V.}},
\oauthor{\bsnm{Lavdas}, \binits{I.}},
\oauthor{\bsnm{Bai}, \binits{W.}},
\oauthor{\bsnm{Kamnitsas}, \binits{K.}},
\oauthor{\bsnm{Aboagye}, \binits{E.O.}},
\oauthor{\bsnm{Rockall}, \binits{A.G.}},
\oauthor{\bsnm{Rueckert}, \binits{D.}},
\oauthor{\bsnm{Glocker}, \binits{B.}}:
{Reverse Classification Accuracy: Predicting Segmentation Performance in the
  Absence of Ground Truth}.
IEEE Transactions on Medical Imaging,
1--1
(2017).
doi:\doiurl{10.1109/TMI.2017.2665165}
\end{botherref}
\endbibitem

\bibitem{Zikic2014}
\begin{barticle}
\bauthor{\bsnm{Zikic}, \binits{D.}},
\bauthor{\bsnm{Glocker}, \binits{B.}},
\bauthor{\bsnm{Criminisi}, \binits{A.}}:
\batitle{{Encoding atlases by randomized classification forests for efficient
  multi-atlas label propagation}}.
\bjtitle{Medical Image Analysis}
\bvolume{18}(\bissue{8}),
\bfpage{1262}--\blpage{1273}
(\byear{2014}).
doi:\doiurl{10.1016/j.media.2014.06.010}
\end{barticle}
\endbibitem

\bibitem{Robinson2017}
\begin{bchapter}
\bauthor{\bsnm{Robinson}, \binits{R.}},
\bauthor{\bsnm{Valindria}, \binits{V.V.}},
\bauthor{\bsnm{Bai}, \binits{W.}},
\bauthor{\bsnm{Suzuki}, \binits{H.}},
\bauthor{\bsnm{Matthews}, \binits{P.M.}},
\bauthor{\bsnm{Page}, \binits{C.}},
\bauthor{\bsnm{Rueckert}, \binits{D.}},
\bauthor{\bsnm{Glocker}, \binits{B.}}:
\bctitle{Automatic quality control of cardiac mri segmentation in large-scale
  population imaging}.
In: \beditor{\bsnm{Descoteaux}, \binits{M.}},
\beditor{\bsnm{Maier-Hein}, \binits{L.}},
\beditor{\bsnm{Franz}, \binits{A.}},
\beditor{\bsnm{Jannin}, \binits{P.}},
\beditor{\bsnm{Collins}, \binits{D.L.}},
\beditor{\bsnm{Duchesne}, \binits{S.}} (eds.)
\bbtitle{Medical Image Computing and Computer Assisted Intervention - MICCAI
  2017},
pp. \bfpage{720}--\blpage{727}.
\bpublisher{Springer},
\blocation{Cham}
(\byear{2017})
\end{bchapter}
\endbibitem

\bibitem{Oktay2017}
\begin{barticle}
\bauthor{\bsnm{Oktay}, \binits{O.}},
\bauthor{\bsnm{Bai}, \binits{W.}},
\bauthor{\bsnm{Guerrero}, \binits{R.}},
\bauthor{\bsnm{Rajchl}, \binits{M.}},
\bauthor{\bparticle{de} \bsnm{Marvao}, \binits{A.}},
\bauthor{\bsnm{O\'Regan}, \binits{D.P.}},
\bauthor{\bsnm{Cook}, \binits{S.A.}},
\bauthor{\bsnm{Heinrich}, \binits{M.P.}},
\bauthor{\bsnm{Glocker}, \binits{B.}},
\bauthor{\bsnm{Rueckert}, \binits{D.}}:
\batitle{Stratified decision forests for accurate anatomical landmark
  localization in cardiac images}.
\bjtitle{{IEEE} Transactions on Medical Imaging}
\bvolume{36}(\bissue{1}),
\bfpage{332}--\blpage{342}
(\byear{2017}).
doi:\doiurl{10.1109/tmi.2016.2597270}
\end{barticle}
\endbibitem

\bibitem{Petersen2016}
\begin{barticle}
\bauthor{\bsnm{Petersen}, \binits{S.E.}},
\bauthor{\bsnm{Matthews}, \binits{P.M.}},
\bauthor{\bsnm{Francis}, \binits{J.M.}},
\bauthor{\bsnm{Robson}, \binits{M.D.}},
\bauthor{\bsnm{Zemrak}, \binits{F.}},
\bauthor{\bsnm{Boubertakh}, \binits{R.}},
\bauthor{\bsnm{Young}, \binits{A.A.}},
\bauthor{\bsnm{Hudson}, \binits{S.}},
\bauthor{\bsnm{Weale}, \binits{P.}},
\bauthor{\bsnm{Garratt}, \binits{S.}},
\bauthor{\bsnm{Collins}, \binits{R.}},
\bauthor{\bsnm{Piechnik}, \binits{S.}},
\bauthor{\bsnm{Neubauer}, \binits{S.}}:
\batitle{Uk biobank's cardiovascular magnetic resonance protocol}.
\bjtitle{Journal of Cardiovascular Magnetic Resonance}
\bvolume{18}(\bissue{1}),
\bfpage{8}
(\byear{2016}).
doi:\doiurl{10.1186/s12968-016-0227-4}
\end{barticle}
\endbibitem

\bibitem{Petersen2017}
\begin{botherref}
\oauthor{\bsnm{Petersen}, \binits{S.E.}},
\oauthor{\bsnm{Aung}, \binits{N.}},
\oauthor{\bsnm{Sanghvi}, \binits{M.M.}},
\oauthor{\bsnm{Zemrak}, \binits{F.}},
\oauthor{\bsnm{Fung}, \binits{K.}},
\oauthor{\bsnm{Paiva}, \binits{J.M.}},
\oauthor{\bsnm{Francis}, \binits{J.M.}},
\oauthor{\bsnm{Khanji}, \binits{M.Y.}},
\oauthor{\bsnm{Lukaschuk}, \binits{E.}},
\oauthor{\bsnm{Lee}, \binits{A.M.}},
\oauthor{\bsnm{Carapella}, \binits{V.}},
\oauthor{\bsnm{Kim}, \binits{Y.J.}},
\oauthor{\bsnm{Leeson}, \binits{P.}},
\oauthor{\bsnm{Piechnik}, \binits{S.K.}},
\oauthor{\bsnm{Neubauer}, \binits{S.}}:
Reference ranges for cardiac structure and function using cardiovascular
  magnetic resonance ({CMR}) in caucasians from the {UK} biobank population
  cohort.
Journal of Cardiovascular Magnetic Resonance
\textbf{19}(1)
(2017).
doi:\doiurl{10.1186/s12968-017-0327-9}
\end{botherref}
\endbibitem

\bibitem{Petersen2017b}
\begin{barticle}
\bauthor{\bsnm{Petersen}, \binits{S.E.}},
\bauthor{\bsnm{Sanghvi}, \binits{M.M.}},
\bauthor{\bsnm{Aung}, \binits{N.}},
\bauthor{\bsnm{Cooper}, \binits{J.A.}},
\bauthor{\bsnm{Paiva}, \binits{J.M.}},
\bauthor{\bsnm{Zemrak}, \binits{F.}},
\bauthor{\bsnm{Fung}, \binits{K.}},
\bauthor{\bsnm{Lukaschuk}, \binits{E.}},
\bauthor{\bsnm{Lee}, \binits{A.M.}},
\bauthor{\bsnm{Carapella}, \binits{V.}},
\bauthor{\bsnm{Kim}, \binits{Y.J.}},
\bauthor{\bsnm{Piechnik}, \binits{S.K.}},
\bauthor{\bsnm{Neubauer}, \binits{S.}}:
\batitle{The impact of cardiovascular risk factors on cardiac structure and
  function: Insights from the uk biobank imaging enhancement study}.
\bjtitle{PLOS ONE}
\bvolume{12}(\bissue{10}),
\bfpage{1}--\blpage{14}
(\byear{2017}).
doi:\doiurl{10.1371/journal.pone.0185114}
\end{barticle}
\endbibitem

\bibitem{Bai2013}
\begin{barticle}
\bauthor{\bsnm{Bai}, \binits{W.}},
\bauthor{\bsnm{Shi}, \binits{W.}},
\bauthor{\bsnm{O`Regan}, \binits{D.P.}},
\bauthor{\bsnm{Tong}, \binits{T.}},
\bauthor{\bsnm{Wang}, \binits{H.}},
\bauthor{\bsnm{Jamil-Copley}, \binits{S.}},
\bauthor{\bsnm{Peters}, \binits{N.S.}},
\bauthor{\bsnm{Rueckert}, \binits{D.}}:
\batitle{{A Probabilistic Patch-Based Label Fusion Model for Multi-Atlas
  Segmentation With Registration Refinement: Application to Cardiac MR
  Images}}.
\bjtitle{IEEE Transactions on Medical Imaging}
\bvolume{32}(\bissue{7}),
\bfpage{1302}--\blpage{1315}
(\byear{2013}).
doi:\doiurl{10.1109/TMI.2013.2256922}
\end{barticle}
\endbibitem

\bibitem{Robinson2018}
\begin{botherref}
\oauthor{\bsnm{{Robinson}}, \binits{R.}},
\oauthor{\bsnm{{Oktay}}, \binits{O.}},
\oauthor{\bsnm{{Bai}}, \binits{W.}},
\oauthor{\bsnm{{Valindria}}, \binits{V.}},
\oauthor{\bsnm{{Sanghvi}}, \binits{M.}},
\oauthor{\bsnm{{Aung}}, \binits{N.}},
\oauthor{\bsnm{{Paiva}}, \binits{J.}},
\oauthor{\bsnm{{Zemrak}}, \binits{F.}},
\oauthor{\bsnm{{Fung}}, \binits{K.}},
\oauthor{\bsnm{{Lukaschuk}}, \binits{E.}},
\oauthor{\bsnm{{Lee}}, \binits{A.}},
\oauthor{\bsnm{{Carapella}}, \binits{V.}},
\oauthor{\bsnm{{Kim}}, \binits{Y.J.}},
\oauthor{\bsnm{{Kainz}}, \binits{B.}},
\oauthor{\bsnm{{Piechnik}}, \binits{S.}},
\oauthor{\bsnm{{Neubauer}}, \binits{S.}},
\oauthor{\bsnm{{Petersen}}, \binits{S.}},
\oauthor{\bsnm{{Page}}, \binits{C.}},
\oauthor{\bsnm{{Rueckert}}, \binits{D.}},
\oauthor{\bsnm{{Glocker}}, \binits{B.}}:
{Real-time Prediction of Segmentation Quality}.
ArXiv e-prints
(2018).
\arxivurl{1806.06244}
\end{botherref}
\endbibitem

\end{thebibliography}

\newcommand{\BMCxmlcomment}[1]{}

\BMCxmlcomment{

<refgrp>

<bibl id="B1">
  <title><p>{UK Biobank: An Open Access Resource for Identifying the Causes of
  a Wide Range of Complex Diseases of Middle and Old Age}</p></title>
  <aug>
    <au><snm>Sudlow</snm><fnm>C</fnm></au>
    <au><snm>Gallacher</snm><fnm>J</fnm></au>
    <au><snm>Allen</snm><fnm>N</fnm></au>
    <au><snm>Beral</snm><fnm>V</fnm></au>
    <au><snm>Burton</snm><fnm>P</fnm></au>
    <au><snm>Danesh</snm><fnm>J</fnm></au>
    <au><snm>Downey</snm><fnm>P</fnm></au>
    <au><snm>Elliott</snm><fnm>P</fnm></au>
    <au><snm>Green</snm><fnm>J</fnm></au>
    <au><snm>Landray</snm><fnm>M</fnm></au>
    <au><snm>Liu</snm><fnm>B</fnm></au>
    <au><snm>Matthews</snm><fnm>P</fnm></au>
    <au><snm>Ong</snm><fnm>G</fnm></au>
    <au><snm>Pell</snm><fnm>J</fnm></au>
    <au><snm>Silman</snm><fnm>A</fnm></au>
    <au><snm>Young</snm><fnm>A</fnm></au>
    <au><snm>Sprosen</snm><fnm>T</fnm></au>
    <au><snm>Peakman</snm><fnm>T</fnm></au>
    <au><snm>Collins</snm><fnm>R</fnm></au>
  </aug>
  <source>PLoS Medicine</source>
  <pubdate>2015</pubdate>
  <volume>12</volume>
  <issue>3</issue>
  <fpage>1</fpage>
  <lpage>-10</lpage>
</bibl>

<bibl id="B2">
  <title><p>{Automated Image Analysis for High-Content Screening and
  Analysis}</p></title>
  <aug>
    <au><snm>Shariff</snm><fnm>A</fnm></au>
    <au><snm>Kangas</snm><fnm>J</fnm></au>
    <au><snm>Coelho</snm><fnm>LP</fnm></au>
    <au><snm>Quinn</snm><fnm>S</fnm></au>
    <au><snm>Murphy</snm><fnm>RF</fnm></au>
  </aug>
  <source>Journal of Biomolecular Screening</source>
  <editor>Trask, O. Joseph and Davies, Anthony and Haney, Steven</editor>
  <pubdate>2010</pubdate>
  <volume>15</volume>
  <issue>7</issue>
  <fpage>726</fpage>
  <lpage>-734</lpage>
  <url>http://journals.sagepub.com/doi/10.1177/1087057110370894</url>
</bibl>

<bibl id="B3">
  <title><p>{Machine learning approaches in medical image analysis: From
  detection to diagnosis.}</p></title>
  <aug>
    <au><snm>Bruijne</snm><fnm>M</fnm></au>
  </aug>
  <source>Medical Image Analysis</source>
  <publisher>Elsevier B.V.</publisher>
  <pubdate>2016</pubdate>
  <volume>33</volume>
  <fpage>94</fpage>
  <lpage>-97</lpage>
</bibl>

<bibl id="B4">
  <title><p>Human-level CMR image analysis with deep fully convolutional
  networks</p></title>
  <aug>
    <au><snm>Bai</snm><fnm>W</fnm></au>
    <au><snm>Sinclair</snm><fnm>M</fnm></au>
    <au><snm>Tarroni</snm><fnm>G</fnm></au>
    <au><snm>Oktay</snm><fnm>O</fnm></au>
    <au><snm>Rajchl</snm><fnm>M</fnm></au>
    <au><snm>Vaillant</snm><fnm>G</fnm></au>
    <au><snm>Lee</snm><fnm>AM</fnm></au>
    <au><snm>Aung</snm><fnm>N</fnm></au>
    <au><snm>Lukaschuk</snm><fnm>E</fnm></au>
    <au><snm>Sanghvi</snm><fnm>MM</fnm></au>
    <au><snm>Zemrak</snm><fnm>F</fnm></au>
    <au><snm>Fung</snm><fnm>K</fnm></au>
    <au><snm>Paiva</snm><fnm>JM</fnm></au>
    <au><snm>Carapella</snm><fnm>V</fnm></au>
    <au><snm>Kim</snm><fnm>YJ</fnm></au>
    <au><snm>Suzuki</snm><fnm>H</fnm></au>
    <au><snm>Kainz</snm><fnm>B</fnm></au>
    <au><snm>Matthews</snm><fnm>PM</fnm></au>
    <au><snm>Petersen</snm><fnm>SE</fnm></au>
    <au><snm>Piechnik</snm><fnm>SK</fnm></au>
    <au><snm>Neubauer</snm><fnm>S</fnm></au>
    <au><snm>Glocker</snm><fnm>B</fnm></au>
    <au><snm>Rueckert</snm><fnm>D</fnm></au>
  </aug>
</bibl>

<bibl id="B5">
  <title><p>{Generalized overlap measures for evaluation and validation in
  medical image analysis}</p></title>
  <aug>
    <au><snm>Crum</snm><fnm>WR</fnm></au>
    <au><snm>Camara</snm><fnm>O</fnm></au>
    <au><snm>Hill</snm><fnm>DLG</fnm></au>
  </aug>
  <source>IEEE Transactions on Medical Imaging</source>
  <pubdate>2006</pubdate>
  <volume>25</volume>
  <issue>11</issue>
  <fpage>1451</fpage>
  <lpage>-1461</lpage>
</bibl>

<bibl id="B6">
  <title><p>{Metrics for evaluating 3D medical image segmentation: analysis,
  selection, and tool.}</p></title>
  <aug>
    <au><snm>Taha</snm><fnm>AA</fnm></au>
    <au><snm>Hanbury</snm><fnm>A</fnm></au>
  </aug>
  <source>BMC medical imaging</source>
  <publisher>BMC Medical Imaging</publisher>
  <pubdate>2015</pubdate>
  <volume>15</volume>
  <fpage>29</fpage>
  <url>http://www.pubmedcentral.nih.gov/articlerender.fcgi?artid=4533825{\&}tool=pmcentrez{\&}rendertype=abstract</url>
</bibl>

<bibl id="B7">
  <title><p>{Towards the Semantic Enrichment of Free-Text Annotation of Image
  Quality Assessment for UK Biobank Cardiac Cine MRI Scans}</p></title>
  <aug>
    <au><snm>Carapella</snm><fnm>V</fnm></au>
    <au><snm>Jim{\'{e}}nez Ruiz</snm><fnm>E</fnm></au>
    <au><snm>Lukaschuk</snm><fnm>E</fnm></au>
    <au><snm>Aung</snm><fnm>N</fnm></au>
    <au><snm>Fung</snm><fnm>K</fnm></au>
    <au><snm>Paiva</snm><fnm>J</fnm></au>
    <au><snm>Sanghvi</snm><fnm>M</fnm></au>
    <au><snm>Neubauer</snm><fnm>S</fnm></au>
    <au><snm>Petersen</snm><fnm>S</fnm></au>
    <au><snm>Horrocks</snm><fnm>I</fnm></au>
    <au><snm>Piechnik</snm><fnm>S</fnm></au>
  </aug>
  <source>MICCAI Workshop on Large-scale Annotation of Biomedical data and
  Expert Label Synthesis (LABELS)</source>
  <publisher>{Cham}: Springer International Publishing</publisher>
  <pubdate>2016</pubdate>
  <fpage>238</fpage>
  <lpage>-248</lpage>
</bibl>

<bibl id="B8">
  <title><p>{Automated Quality Assessment of Cardiac MR Images Using
  Convolutional Neural Networks}</p></title>
  <aug>
    <au><snm>Zhang</snm><fnm>L</fnm></au>
    <au><snm>Gooya</snm><fnm>A</fnm></au>
    <au><snm>Dong</snm><fnm>B</fnm></au>
    <au><snm>Hua</snm><fnm>R</fnm></au>
    <au><snm>Petersen</snm><fnm>SE</fnm></au>
    <au><snm>Medrano Gracia</snm><fnm>P</fnm></au>
    <au><snm>Frangi</snm><fnm>AF</fnm></au>
  </aug>
  <source>Medical Image Computing and Computer-Assisted Intervention –
  SASHIMI 2016</source>
  <publisher>{Cham}: Springer International Publishing</publisher>
  <editor>Tsaftaris, Sotirios A. and Gooya, Ali and Frangi, Alejandro F. and
  Prince, Jerry L.</editor>
  <series><title><p>Lecture Notes in Computer Science</p></title></series>
  <pubdate>2016</pubdate>
  <volume>9968</volume>
  <fpage>138</fpage>
  <lpage>-145</lpage>
</bibl>

<bibl id="B9">
  <title><p>{Cross Validation Framework to Choose amongst Models and Datasets
  for Transfer Learning}</p></title>
  <aug>
    <au><snm>Zhong</snm><fnm>E</fnm></au>
    <au><snm>Fan</snm><fnm>W</fnm></au>
    <au><snm>Yang</snm><fnm>Q</fnm></au>
    <au><snm>Verscheure</snm><fnm>O</fnm></au>
    <au><snm>Ren</snm><fnm>J</fnm></au>
  </aug>
  <source>Lecture Notes in Computer Science (including subseries Lecture Notes
  in Artificial Intelligence and Lecture Notes in Bioinformatics)</source>
  <publisher>Springer Berlin Heidelberg</publisher>
  <pubdate>2010</pubdate>
  <volume>6323 LNAI</volume>
  <issue>PART 3</issue>
  <fpage>547</fpage>
  <lpage>-562</lpage>
  <url>http://link.springer.com/10.1007/978-3-642-15939-8{\_}35</url>
</bibl>

<bibl id="B10">
  <title><p>{Reverse testing}</p></title>
  <aug>
    <au><snm>Fan</snm><fnm>W</fnm></au>
    <au><snm>Davidson</snm><fnm>I</fnm></au>
  </aug>
  <source>Proceedings of the 12th ACM SIGKDD international conference on
  Knowledge discovery and data mining - KDD '06</source>
  <publisher>New York, New York, USA: ACM Press</publisher>
  <pubdate>2006</pubdate>
  <fpage>147</fpage>
</bibl>

<bibl id="B11">
  <title><p>{Reverse Classification Accuracy: Predicting Segmentation
  Performance in the Absence of Ground Truth}</p></title>
  <aug>
    <au><snm>Valindria</snm><fnm>VV</fnm></au>
    <au><snm>Lavdas</snm><fnm>I</fnm></au>
    <au><snm>Bai</snm><fnm>W</fnm></au>
    <au><snm>Kamnitsas</snm><fnm>K</fnm></au>
    <au><snm>Aboagye</snm><fnm>EO</fnm></au>
    <au><snm>Rockall</snm><fnm>AG</fnm></au>
    <au><snm>Rueckert</snm><fnm>D</fnm></au>
    <au><snm>Glocker</snm><fnm>B</fnm></au>
  </aug>
  <source>IEEE Transactions on Medical Imaging</source>
  <pubdate>2017</pubdate>
  <fpage>1</fpage>
  <lpage>-1</lpage>
  <url>http://ieeexplore.ieee.org/document/7902121/</url>
</bibl>

<bibl id="B12">
  <title><p>{Encoding atlases by randomized classification forests for
  efficient multi-atlas label propagation}</p></title>
  <aug>
    <au><snm>Zikic</snm><fnm>D.</fnm></au>
    <au><snm>Glocker</snm><fnm>B.</fnm></au>
    <au><snm>Criminisi</snm><fnm>A.</fnm></au>
  </aug>
  <source>Medical Image Analysis</source>
  <pubdate>2014</pubdate>
  <volume>18</volume>
  <issue>8</issue>
  <fpage>1262</fpage>
  <lpage>-1273</lpage>
  <url>http://linkinghub.elsevier.com/retrieve/pii/S1361841514001042</url>
</bibl>

<bibl id="B13">
  <title><p>Automatic Quality Control of Cardiac MRI Segmentation in
  Large-Scale Population Imaging</p></title>
  <aug>
    <au><snm>Robinson</snm><fnm>R</fnm></au>
    <au><snm>Valindria</snm><fnm>VV</fnm></au>
    <au><snm>Bai</snm><fnm>W</fnm></au>
    <au><snm>Suzuki</snm><fnm>H</fnm></au>
    <au><snm>Matthews</snm><fnm>PM</fnm></au>
    <au><snm>Page</snm><fnm>C</fnm></au>
    <au><snm>Rueckert</snm><fnm>D</fnm></au>
    <au><snm>Glocker</snm><fnm>B</fnm></au>
  </aug>
  <source>Medical Image Computing and Computer Assisted Intervention - MICCAI
  2017</source>
  <publisher>Cham: Springer International Publishing</publisher>
  <editor>Descoteaux, Maxime and Maier-Hein, Lena and Franz, Alfred and Jannin,
  Pierre and Collins, D. Louis and Duchesne, Simon</editor>
  <pubdate>2017</pubdate>
  <fpage>720</fpage>
  <lpage>-727</lpage>
</bibl>

<bibl id="B14">
  <title><p>Stratified Decision Forests for Accurate Anatomical Landmark
  Localization in Cardiac Images</p></title>
  <aug>
    <au><snm>Oktay</snm><fnm>O</fnm></au>
    <au><snm>Bai</snm><fnm>W</fnm></au>
    <au><snm>Guerrero</snm><fnm>R</fnm></au>
    <au><snm>Rajchl</snm><fnm>M</fnm></au>
    <au><snm>Marvao</snm><fnm>A</fnm></au>
    <au><snm>O\'Regan</snm><fnm>DP</fnm></au>
    <au><snm>Cook</snm><fnm>SA</fnm></au>
    <au><snm>Heinrich</snm><fnm>MP</fnm></au>
    <au><snm>Glocker</snm><fnm>B</fnm></au>
    <au><snm>Rueckert</snm><fnm>D</fnm></au>
  </aug>
  <source>{IEEE} Transactions on Medical Imaging</source>
  <publisher>Institute of Electrical and Electronics Engineers
  ({IEEE})</publisher>
  <pubdate>2017</pubdate>
  <volume>36</volume>
  <issue>1</issue>
  <fpage>332</fpage>
  <lpage>-342</lpage>
</bibl>

<bibl id="B15">
  <title><p>UK Biobank's cardiovascular magnetic resonance protocol</p></title>
  <aug>
    <au><snm>Petersen</snm><fnm>SE</fnm></au>
    <au><snm>Matthews</snm><fnm>PM</fnm></au>
    <au><snm>Francis</snm><fnm>JM</fnm></au>
    <au><snm>Robson</snm><fnm>MD</fnm></au>
    <au><snm>Zemrak</snm><fnm>F</fnm></au>
    <au><snm>Boubertakh</snm><fnm>R</fnm></au>
    <au><snm>Young</snm><fnm>AA</fnm></au>
    <au><snm>Hudson</snm><fnm>S</fnm></au>
    <au><snm>Weale</snm><fnm>P</fnm></au>
    <au><snm>Garratt</snm><fnm>S</fnm></au>
    <au><snm>Collins</snm><fnm>R</fnm></au>
    <au><snm>Piechnik</snm><fnm>S</fnm></au>
    <au><snm>Neubauer</snm><fnm>S</fnm></au>
  </aug>
  <source>Journal of Cardiovascular Magnetic Resonance</source>
  <pubdate>2016</pubdate>
  <volume>18</volume>
  <issue>1</issue>
  <fpage>8</fpage>
  <url>https://doi.org/10.1186/s12968-016-0227-4</url>
</bibl>

<bibl id="B16">
  <title><p>Reference ranges for cardiac structure and function using
  cardiovascular magnetic resonance ({CMR}) in Caucasians from the {UK} Biobank
  population cohort</p></title>
  <aug>
    <au><snm>Petersen</snm><fnm>SE</fnm></au>
    <au><snm>Aung</snm><fnm>N</fnm></au>
    <au><snm>Sanghvi</snm><fnm>MM</fnm></au>
    <au><snm>Zemrak</snm><fnm>F</fnm></au>
    <au><snm>Fung</snm><fnm>K</fnm></au>
    <au><snm>Paiva</snm><fnm>JM</fnm></au>
    <au><snm>Francis</snm><fnm>JM</fnm></au>
    <au><snm>Khanji</snm><fnm>MY</fnm></au>
    <au><snm>Lukaschuk</snm><fnm>E</fnm></au>
    <au><snm>Lee</snm><fnm>AM</fnm></au>
    <au><snm>Carapella</snm><fnm>V</fnm></au>
    <au><snm>Kim</snm><fnm>YJ</fnm></au>
    <au><snm>Leeson</snm><fnm>P</fnm></au>
    <au><snm>Piechnik</snm><fnm>SK</fnm></au>
    <au><snm>Neubauer</snm><fnm>S</fnm></au>
  </aug>
  <source>Journal of Cardiovascular Magnetic Resonance</source>
  <publisher>Springer Nature</publisher>
  <pubdate>2017</pubdate>
  <volume>19</volume>
  <issue>1</issue>
</bibl>

<bibl id="B17">
  <title><p>The impact of cardiovascular risk factors on cardiac structure and
  function: Insights from the UK Biobank imaging enhancement study</p></title>
  <aug>
    <au><snm>Petersen</snm><fnm>SE</fnm></au>
    <au><snm>Sanghvi</snm><fnm>MM</fnm></au>
    <au><snm>Aung</snm><fnm>N</fnm></au>
    <au><snm>Cooper</snm><fnm>JA</fnm></au>
    <au><snm>Paiva</snm><fnm>JM</fnm></au>
    <au><snm>Zemrak</snm><fnm>F</fnm></au>
    <au><snm>Fung</snm><fnm>K</fnm></au>
    <au><snm>Lukaschuk</snm><fnm>E</fnm></au>
    <au><snm>Lee</snm><fnm>AM</fnm></au>
    <au><snm>Carapella</snm><fnm>V</fnm></au>
    <au><snm>Kim</snm><fnm>YJ</fnm></au>
    <au><snm>Piechnik</snm><fnm>SK</fnm></au>
    <au><snm>Neubauer</snm><fnm>S</fnm></au>
  </aug>
  <source>PLOS ONE</source>
  <publisher>Public Library of Science</publisher>
  <pubdate>2017</pubdate>
  <volume>12</volume>
  <issue>10</issue>
  <fpage>1</fpage>
  <lpage>14</lpage>
  <url>https://doi.org/10.1371/journal.pone.0185114</url>
</bibl>

<bibl id="B18">
  <title><p>{A Probabilistic Patch-Based Label Fusion Model for Multi-Atlas
  Segmentation With Registration Refinement: Application to Cardiac MR
  Images}</p></title>
  <aug>
    <au><snm>Bai</snm><fnm>W</fnm></au>
    <au><snm>Shi</snm><fnm>W</fnm></au>
    <au><snm>O`Regan</snm><fnm>DP</fnm></au>
    <au><snm>Tong</snm><fnm>T</fnm></au>
    <au><snm>Wang</snm><fnm>H</fnm></au>
    <au><snm>Jamil Copley</snm><fnm>S</fnm></au>
    <au><snm>Peters</snm><fnm>NS</fnm></au>
    <au><snm>Rueckert</snm><fnm>D</fnm></au>
  </aug>
  <source>IEEE Transactions on Medical Imaging</source>
  <pubdate>2013</pubdate>
  <volume>32</volume>
  <issue>7</issue>
  <fpage>1302</fpage>
  <lpage>-1315</lpage>
  <url>http://ieeexplore.ieee.org/document/6494647/</url>
</bibl>

<bibl id="B19">
  <title><p>{Real-time Prediction of Segmentation Quality}</p></title>
  <aug>
    <au><snm>{Robinson}</snm><fnm>R.</fnm></au>
    <au><snm>{Oktay}</snm><fnm>O.</fnm></au>
    <au><snm>{Bai}</snm><fnm>W.</fnm></au>
    <au><snm>{Valindria}</snm><fnm>V.</fnm></au>
    <au><snm>{Sanghvi}</snm><fnm>M.</fnm></au>
    <au><snm>{Aung}</snm><fnm>N.</fnm></au>
    <au><snm>{Paiva}</snm><fnm>J.</fnm></au>
    <au><snm>{Zemrak}</snm><fnm>F.</fnm></au>
    <au><snm>{Fung}</snm><fnm>K.</fnm></au>
    <au><snm>{Lukaschuk}</snm><fnm>E.</fnm></au>
    <au><snm>{Lee}</snm><fnm>A.</fnm></au>
    <au><snm>{Carapella}</snm><fnm>V.</fnm></au>
    <au><snm>{Kim}</snm><fnm>Y. J.</fnm></au>
    <au><snm>{Kainz}</snm><fnm>B.</fnm></au>
    <au><snm>{Piechnik}</snm><fnm>S.</fnm></au>
    <au><snm>{Neubauer}</snm><fnm>S.</fnm></au>
    <au><snm>{Petersen}</snm><fnm>S.</fnm></au>
    <au><snm>{Page}</snm><fnm>C.</fnm></au>
    <au><snm>{Rueckert}</snm><fnm>D.</fnm></au>
    <au><snm>{Glocker}</snm><fnm>B.</fnm></au>
  </aug>
  <source>ArXiv e-prints</source>
  <pubdate>2018</pubdate>
</bibl>

</refgrp>
} 






\end{backmatter}

\end{document}